\documentclass{article} 
\usepackage{iclr2027_conference,times}

\usepackage{amsmath,amsfonts,bm}

\def\eqref#1{equation~\ref{#1}}

\def\1{\bm{1}}

\DeclareMathAlphabet{\mathsfit}{\encodingdefault}{\sfdefault}{m}{sl}
\SetMathAlphabet{\mathsfit}{bold}{\encodingdefault}{\sfdefault}{bx}{n}

\usepackage{hyperref}
\usepackage{url}
\usepackage{amssymb}
\usepackage{booktabs}
\usepackage{tabularx}
\usepackage{array}
\usepackage{graphicx}
\usepackage[table]{xcolor}
\usepackage{mdframed}
\usepackage{listings}
\usepackage{xcolor}
\usepackage{subcaption}
\usepackage{pifont}
\usepackage{algorithm}
\usepackage{algpseudocode}
\usepackage{multirow}
\usepackage{enumitem}
\usepackage{capt-of}

\usepackage{bbm}

\lstdefinestyle{promptstyle}{
    basicstyle=\ttfamily\scriptsize,
    columns=fullflexible,
    keepspaces=true,
    breaklines=true,
    breakatwhitespace=false,
    showstringspaces=false,
    numbers=none,
    tabsize=2
}

\usepackage{etoc}
\usepackage{wrapfig}

\title{\method{}: Grounding Skill Evolution in Replayable Evidence}

\author{%
\parbox[t]{\textwidth}{%
\raggedright
\normalfont\normalsize
\textbf{%
Yan Zhou\textsuperscript{1*},
Yili Wang\textsuperscript{1*},
Yiwei Dai\textsuperscript{1},
Qinggang Zhang\textsuperscript{1\(\dagger\)}, 
Xin Wang\textsuperscript{1\(\dagger\)}
}\\[2pt]
\textsuperscript{1}School of Artificial Intelligence,
Jilin University, Changchun, China\\
\texttt{\{zhouyan25,daiyw25\}@mails.jlu.edu.cn}\\
\texttt{\{wangyili,qinggangzhang,xinwang\}@jlu.edu.cn}
}%
}

\newcommand{\method}{\textsc{EviSkill}}

\iclrfinalcopy 
\begin{document}

\maketitle

\lhead{Preprint}
\begingroup
\renewcommand{\thefootnote}{\fnsymbol{footnote}}
\footnotetext[1]{Equal contribution. \textsuperscript{\(\dagger\)}Corresponding author.}
\endgroup

\begin{abstract}
Continual skill evolution enables LLM agents to accumulate and refine reusable procedural knowledge from interaction experience without updating model parameters. Its effectiveness depends on determining not only what to change, but also why a change is justified and when it should become persistent guidance. However, existing experience-driven methods can lose the behavioral evidence and task contexts supporting edits. Moreover, a global validation outcome provides an incomplete judgment of its constituent changes: locally supported corrections may be discarded with a rejected revision, while evidence may require further experience to inform useful updates. To this end, we introduce \method{}, an evidence-driven framework that organizes execution observations into Replayable Evidence Cards and synthesizes edits with explicit links to their supporting contexts. Targeted replay verifies these edits through re-execution and provides feedback for correction. Across epochs, \method{} preserves evidence and provisionally retains supported edits for further refinement, while global validation governs their incorporation into the final skill. Experiments on three interactive benchmarks across six LLM backbones demonstrate the effectiveness of this approach. Our code and implementation details are available at \textcolor{blue}{\url{https://github.com/Zhouyaner/eviskill}}.
\end{abstract}
\vspace{-10pt}
\etocdepthtag.toc{main}

\section{Introduction}


LLM agents are increasingly used to solve complex interactive tasks involving tool use, web navigation, and long-horizon decision making~\citep{react,AWM,webarena,mind2web}. These tasks often require reusable procedures for choosing tools, ordering actions, and satisfying domain-specific constraints~\citep{toolllm}. External skills provide such guidance as inspectable artifacts containing procedural instructions and supporting resources that agents can use at inference time without modifying model parameters~\citep{voyager,skillnet2026,feng2026graph}. Empirical studies show that curated skills can improve task completion across diverse domains, while generated and refined skills can yield further gains~\citep{skillsbench,skillgen2026,skillaudit2026,skillaxe2026}. Sustaining these benefits depends on keeping skills effective as agents encounter new tasks and execution conditions.

Recent work uses agent trajectories to create and revise reusable skills.~\citep{autoskill2026,skillevolver,skillr12026} Trace2Skill extracts lessons from individual trajectories and consolidates them into skill patches~\citep{trace2skill2026}, while SkillDisCo compiles procedures shared across successful traces~\citep{skilldisco}. SkillGrad converts execution diagnoses into textual update signals for editing skill packages~\citep{skillgrad2026}. SkillAdaptor locates an actionable failure step and revises the skill instruction implicated in that failure~\citep{skilladaptor2026}. SkillRevise repairs an initially LLM-authored skill using feedback from its execution traces~\citep{skillrevise2026}. EvoSkill proposes new skills or revises existing ones through failure analysis~\citep{evoskill2026}. SkillOpt proposes bounded edits to a skill document and accepts updates that improve a held-out validation score~\citep{skillopt2026}. These studies establish automated skill construction and iterative refinement as practical approaches to improving agent~\citep{xiang2026systematic}.

\begin{wrapfigure}{l}{0.45\textwidth}
    \vspace{-\intextsep}
    \centering
    \includegraphics[width=\linewidth]{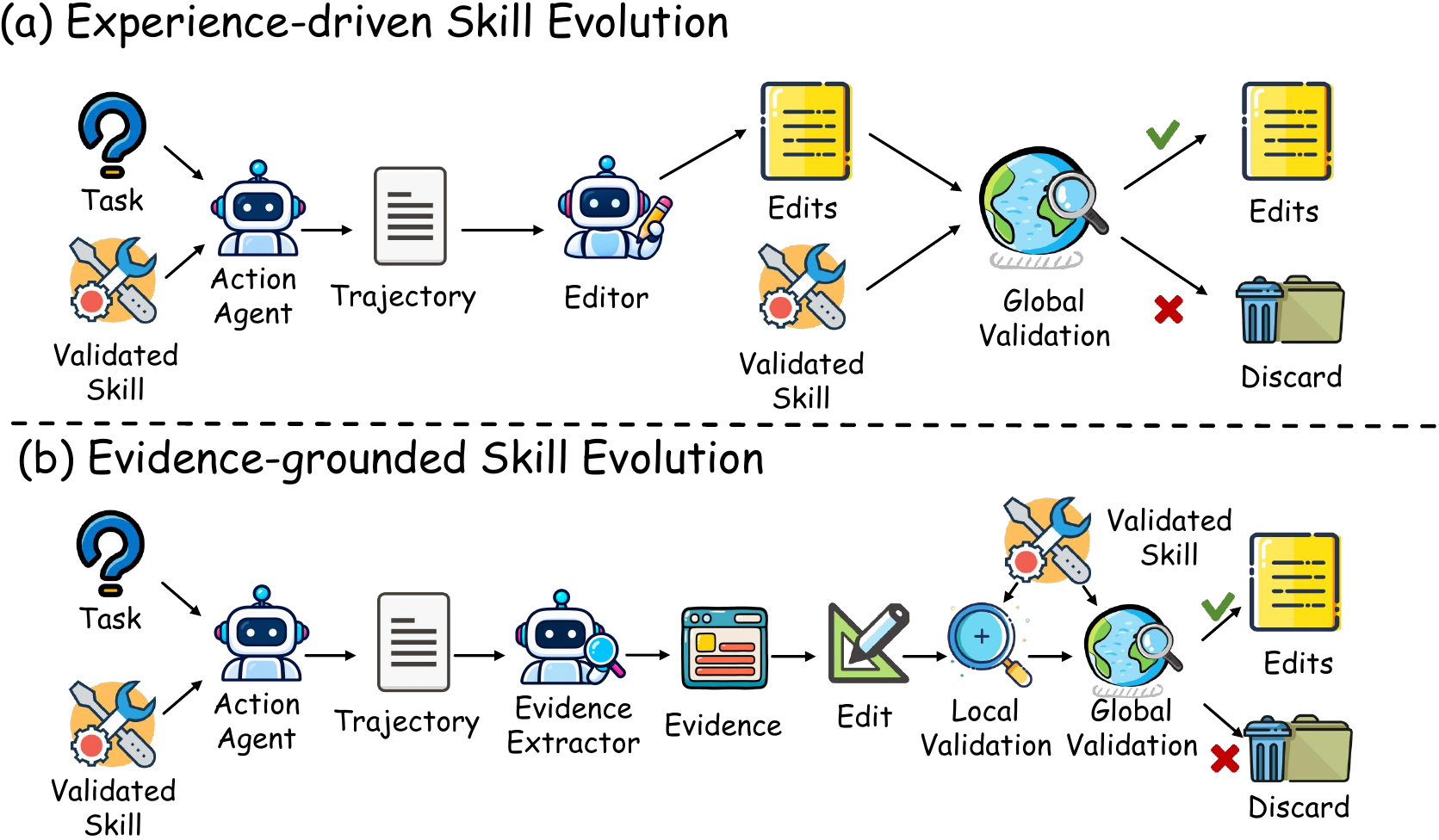}
    \vspace{-21pt}
    \caption{Experience-driven Skill Evolution vs Evidence-grounded Skill Evolution.}
    \label{fig:motivation}
    \vspace{-15pt}
\end{wrapfigure}

Despite this progress, existing skill evolution methods mainly follow an \emph{experience-driven paradigm} in which execution feedback is transformed into reusable guidance and skills are iteratively edited. However, execution experience is inherently local and context-dependent, whereas skill updates are intended to persist and generalize across future interactions. A correction induced from a failed interaction may therefore overfit the observed case, while a procedure distilled from successful execution may capture incidental rather than reusable behavior. Once incorporated, such errors can repeatedly affect future executions and further shape subsequent rounds of skill evolution. These limitations motivate \emph{evidence-driven skill evolution}, which explicitly preserves the execution basis of each proposed edit, examines whether the edited skill produces the intended behavioral change, and revisits that support as additional experience accumulates (Fig.~\ref{fig:motivation}). Under this view, skill evolution is not merely the accumulation of experience-derived edits, but a process of establishing, testing, and revising the support for each change before it is committed as reusable knowledge.

Evidence-based skill evolution is promising but challenging: (i) \textbf{Extracted evidence should be reusable and replayable}. Raw trajectories contain task-specific context, informative behavioral signals, redundant actions, and incidental interactions. Simply preserving complete trajectories makes the relevant support difficult to isolate, whereas aggressively abstracting them into textual lessons may remove the execution context needed for later examination. The challenge is therefore to extract compact, reusable evidence while retaining a traceable connection to the behaviors that justify the proposed change.
(ii) \textbf{Evidence must remain revisable over time:} A locally supported edit may later expose missing conditions, while an initially uncertain correction may become better supported as additional interactions accumulate. 
Continual skill evolution therefore requires a mechanism to repeatedly validate and update evolutionary information before committing it to the reusable skill.

To address this challenge, we introduce \method{}, an \emph{evidence-driven framework} that organizes continual skill evolution into evidence construction, behavioral verification, and cross-epoch refinement. Specifically, rather than directly converting trajectory feedback into committed skill changes, \method{} organizes continual skill evolution into three complementary stages. First, \textbf{Evidence-Grounded Edit Synthesis} transforms localized execution observations into \emph{Replayable Evidence Cards} and constructs candidate edits while preserving explicit links to the task contexts and trajectory ranges that justify them. Second, \textbf{Replay-Guided Edit Verification} re-executes the referenced behaviors under the edited skill, testing whether each candidate actually produces its \emph{intended behavioral effect} rather than relying on trajectory-derived plausibility alone. Third, \textbf{Cross-Epoch Evidence Propagation} maintains evolutionary information \emph{beyond individual update decisions}. Instead of treating immediate outcomes as final judgments, it allows previously generated evidence and revisions to be repeatedly validated, refined, and reconsidered as new experience accumulates. These components make skill evolution evidence-grounded at construction, behaviorally tested at verification, and progressively adjudicated across epochs before knowledge is committed to the reusable skill. Our main contributions are summarized as follows:
\begin{itemize}[leftmargin=*, itemsep=1pt, parsep=0pt, topsep=2pt]

\item
We identify the key limitations of existing experience-driven skill evolution: execution evidence motivating a skill revision is not explicitly maintained with the resulting revision, and evolutionary information generated in one round cannot be continuously reconsidered across epochs. 

\item
we introduce \method{}, an evidence-driven skill evolution framework that organizes continual skill evolution into evidence construction, behavioral verification, and cross-epoch refinement. 

\item
We evaluate \method{} on three real-world benchmarks across multiple LLM backbones, demonstrating consistent improvements over strong skill evolution baselines. Model analyses and case studies show that evidence-grounded replay filters unsupported revisions, while cross-epoch propagation enables prior revisions and evidence to support continued refinement.




\end{itemize}

\vspace{-10pt}
\section{Preliminary Study}
\label{sec:preliminary}

In this section, we conduct two preliminary studies to investigate the reliability of existing skill revision processes in experience-driven skill evolution across three interactive benchmark datasets. Our findings reveal two critical limitations: experience-derived edits do not consistently produce their intended effects in execution, and a candidate revision rejected by global validation may still contain effective constituent edits whose benefits are offset by other edits in the same revision.

\subsection{Behavioral Outcomes of Experience-Derived Edits.}
\label{sec:prelim-outcome}

\begin{wrapfigure}{r}{0.36\textwidth}
\vspace{-8pt}
    \vspace{-\intextsep}
    \centering
    \includegraphics[width=\linewidth]{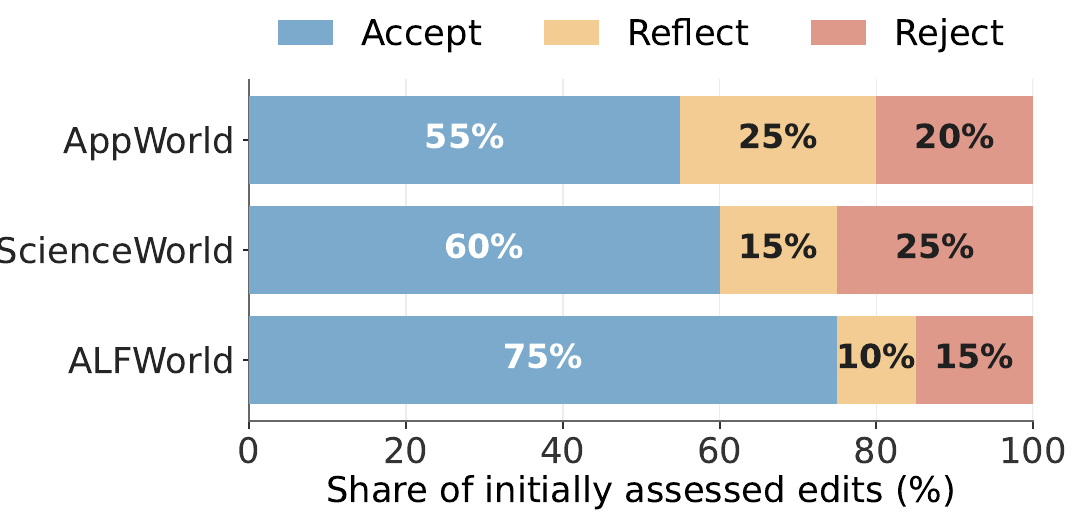}
    \vspace{-21pt}
    \caption{Re-execution outcomes of Experience-Derived Edits.}
    \label{fig:preliminary-a}
    \vspace{-15pt}
\end{wrapfigure}
To examine whether edits generated from execution trajectories improve the behaviors that motivate them, we apply edits generated by Trace2Skill~\citep{trace2skill2026} to the original skill and re-execute their associated trajectory tasks. As shown in Figure~\ref{fig:preliminary-a}, these edits exhibit diverse outcomes across all benchmarks. While some improve task performance, others produce no improvement or even degrade the original behaviors. This observation indicates that deriving edits from execution trajectories does not guarantee behavioral improvement, motivating explicit associations between edits and their supporting execution evidence so that their effects can be directly verified.

\subsection{Useful Information within Rejected Revisions}
\label{sec:prelim-rejected}

\begin{wrapfigure}{l}{0.46\textwidth}
    \vspace{-\intextsep}
    \centering
    \includegraphics[width=\linewidth]{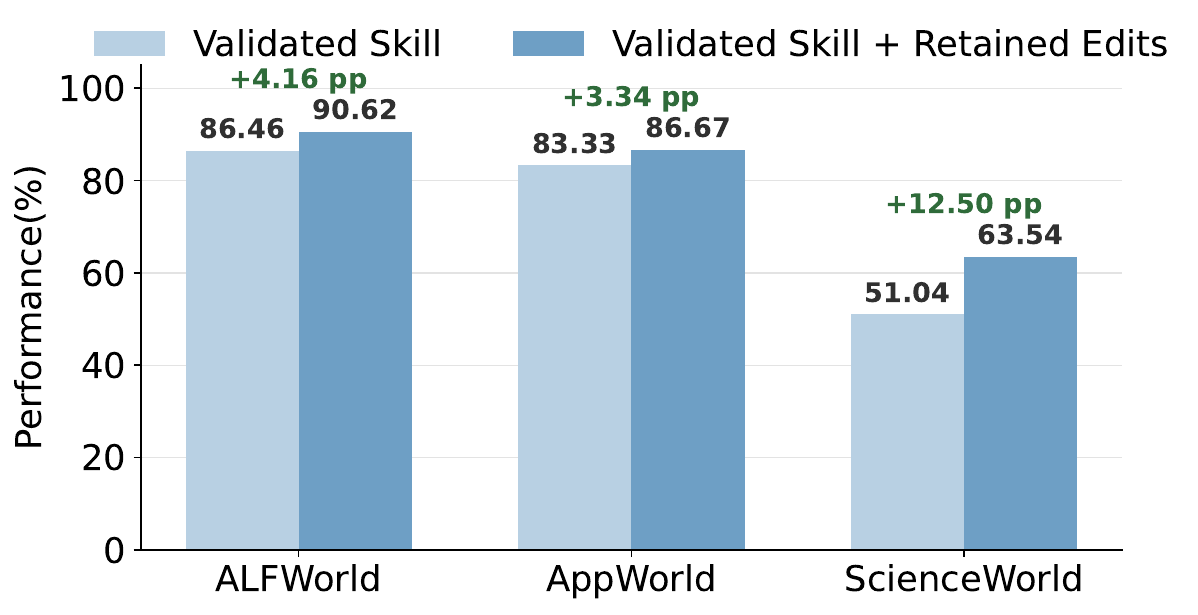}
    \vspace{-21pt}
    \caption{Performance of selectively retained edits from globally rejected revisions.}
    \label{fig:preliminary-b}
    \vspace{-15pt}
\end{wrapfigure}
We further investigate the information contained in unsuccessful evolution rounds. Existing evolution processes typically discard a complete revision when it fails to improve the current skill. However, a revision may consist of multiple edits with different effects, and the failure of the overall revision does not necessarily indicate that every constituent edit is ineffective. As shown in Figure~\ref{fig:preliminary-b}, selectively retaining edits from globally rejected revisions can improve performance on the associated trajectory tasks compared with the previous validated skill. This observation suggests that unsuccessful evolution rounds may still contain valuable information that can contribute to future refinement. These observations show that \textbf{ an edit is not necessarily effective because it is derived from execution experience, nor ineffective because the complete revision containing it is rejected.} This motivates our evidence-grounded framework, which explicitly associates edits with their supporting evidence, verifies their behavioral effects through replay, and preserves evolutionary information across epochs for further validation and refinement.



\vspace{-5pt}
\section{Method}
\label{sec:method}

\begin{figure*}[t]
    \centering
    \includegraphics[width=\textwidth]{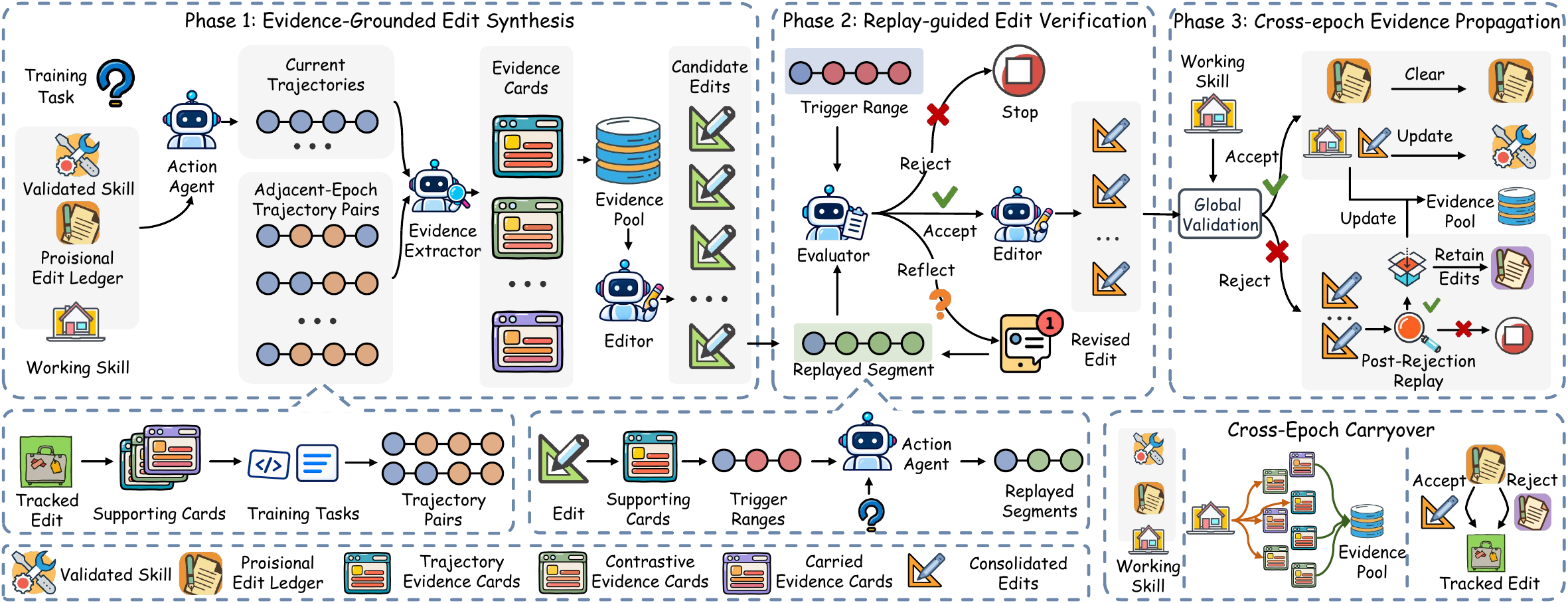}
    \caption{Overview of \method{}. Current trajectories and cross-epoch comparisons provide evidence for edit synthesis. Replay verifies and refines edits, while global validation and post-rejection replay govern skill updates, provisional edit retention, and evidence reuse across epochs.}
    \label{fig:framework}
    \vspace{-10pt}
\end{figure*}

As illustrated in Figure~\ref{fig:framework}, \method{} organizes skill evolution around Replayable Evidence Cards, which record proposed skill corrections and their supporting execution evidence. \textbf{Evidence-Grounded Edit Synthesis} extracts these Cards from trajectories and generates localized edits, each linked to the Cards that motivate it. \textbf{Replay-Guided Edit Verification} uses these links to reconstruct execution contexts and assess or refine edits through replay. \textbf{Cross-Epoch Evidence Propagation} updates the Validated Skill through global validation, retains replay-supported edits provisionally for further training, and preserves evidence for subsequent synthesis and correction.

\vspace{-5pt}
\subsection{Evidence-Grounded Edit Synthesis}
\label{sec:edit_construction}

Each evolution epoch begins with training interactions that provide evidence for skill revision. We distinguish the \emph{Validated Skill}, updated only when a candidate revision improves validation performance, from the \emph{Working Skill} used for these interactions and further revision. The latter incorporates provisionally retained edits so their effects can be examined across epochs.

\noindent\textbf{Definition 1 (Working Skill).} \textit{At epoch $e$, the Working Skill $S_e^W$ combines the latest Validated Skill $S_e^V$ with the Provisional Edit Ledger $P_e$, an ordered collection of replay-supported edits not yet incorporated into $S_e^V$ but retained for further validation and refinement across epochs:}

\begin{equation}
S_e^W=S_e^V\oplus P_e,
\label{eq:working_skill}
\end{equation}
\textit{where $\oplus$ applies the edits in order, and each ledger entry retains references to its supporting execution evidence.}
Given an initial skill $S_0$, we initialize $S_1^V=S_0$ and $P_1=\varnothing$, yielding $S_1^W=S_0$. Section~\ref{sec:evidence_propagation} describes how validation and replay update these states at each epoch.

Let $\mathcal{D}_{\mathrm{tr}}$, $\mathcal{D}_{\mathrm{val}}$, and $\mathcal{D}_{\mathrm{test}}$ denote disjoint training, validation, and test task sets, respectively. At epoch $e$, the Action Agent $\pi$ collects one trajectory $\tau_x^e\sim\pi(\cdot\mid x,S_e^W)$ for each task $x\in\mathcal{D}_{\mathrm{tr}}$, yielding the trajectory set $\mathcal{R}_e$. These trajectories, including the agent's actions, observations, and task outcomes, provide evidence for new edits and for reassessing edits already used by the Working Skill.

\textbf{Evidence extraction and grounding.} An LLM evidence extractor $f^{\mathrm{ext}}$ analyzes the Working Skill $S_e^W$ and its trajectories $\mathcal{R}_e$ to propose corrections or improvements to procedures, preconditions, and recovery guidance. Each proposal is recorded with its supporting trajectory intervals as a Replayable Evidence Card $C_i$ for evidence-grounded edit synthesis and later replay-guided edit verification.

\noindent\textbf{Definition 2 (Replayable Evidence Card).} \textit{A Replayable Evidence Card $C_i$, hereafter a card, records a proposed skill correction and its supporting execution evidence:}
\begin{equation}
C_i=\langle i,\mathcal{G}_i,d_i\rangle,
\label{eq:evidence_card}
\end{equation}
\textit{where $i$ is a persistent identifier for tracking evidence across epochs, $\mathcal{G}_i$ is a set of supporting trigger ranges from trajectories, and $d_i$ is the proposed correction. Each trigger range $g=(x,e_g,s,t)$ identifies the interval $[s,t]$ in the source trajectory $\tau_x^{e_g}$ of task $x$ at source epoch $e_g$. A card may include multiple ranges from the same or different tasks when they support the same correction.}

All cards share this representation. The sets introduced below distinguish their extraction sources or cross-epoch retention. Within each $C_i$, $d_i$ guides edit synthesis and $\mathcal{G}_i$ locates execution contexts for replay. For an edit $u$, its \emph{supporting cards} $\mathcal{C}(u)$ form a set of such cards, linked through persistent identifiers to preserve access to the original execution evidence underlying each edit across epochs.

We extract \emph{Trajectory Evidence Cards} $\mathcal{C}_e^{\mathrm{traj}}$ from current trajectories as $\mathcal{C}_e^{\mathrm{traj}}=f^{\mathrm{ext}}(S_e^W,\mathcal{R}_e)$. To reassess existing edits, we also extract \emph{Contrastive Evidence Cards} $\mathcal{C}_e^{\mathrm{con}}$ from trajectory comparisons across adjacent epochs for training tasks identified by the edits' supporting cards.


These comparisons focus on the \emph{Tracked Edits} $\mathcal{Q}_e$, comprising the live provisional edits in $P_e$ retained through post-rejection replay in prior epochs and the edits newly incorporated into $S_e^V$ during the preceding epoch transition, with $\mathcal{Q}_1=\varnothing$. For each edit $u$, let $\mathcal{C}(u)$ denote the set of cards linked to it as supporting evidence. For $e>1$, these cards for $u\in\mathcal{Q}_e$ identify the training tasks whose complete trajectory pairs $(\tau_x^{e-1},\tau_x^e)$ form the comparison set $\mathcal{R}_e^{\mathrm{cmp}}$. The extractor uses these pairs and the linked edits and cards to identify persistent deficiencies or newly exposed problems:
\begin{equation}
\mathcal{C}_e^{\mathrm{con}}
=f^{\mathrm{ext}}\!\left(S_e^W,\{(u,\mathcal{C}(u)):u\in\mathcal{Q}_e\},\mathcal{R}_e^{\mathrm{cmp}}\right).
\label{eq:contrastive_cards}
\end{equation}
Each $C_i\in\mathcal{C}_e^{\mathrm{con}}$ follows Definition~2 and, when targeting an existing edit, retains that association as metadata for subsequent synthesis and targeted correction of that edit.
The \emph{Evidence Pool} $\mathcal{C}_e$ combines both sources with the carried evidence cards $\mathcal{C}_e^{\mathrm{carry}}$ preserved from preceding epochs:
\begin{equation}
\mathcal{C}_e=\mathcal{C}_e^{\mathrm{carry}}\cup\mathcal{C}_e^{\mathrm{traj}}\cup\mathcal{C}_e^{\mathrm{con}}.
\label{eq:evidence_propagation}
\end{equation}
Cards in $\mathcal{C}_e^{\mathrm{carry}}$ may originate from either source, with their links to the supporting trajectory ranges preserved. Their lifecycle states determine eligibility for new edit synthesis or retention for reference, as described in Section~\ref{sec:evidence_propagation}. Initially, $\mathcal{C}_1^{\mathrm{carry}}=\mathcal{C}_1^{\mathrm{con}}=\varnothing$, yielding $\mathcal{C}_1=\mathcal{C}_1^{\mathrm{traj}}$.

\textbf{Evidence grouping and edit synthesis.} Active cards remain eligible for edit synthesis. We group those with compatible source contexts and correction intents into \emph{Evidence Windows} $\mathcal{B}_k\subseteq\mathcal{C}_e$, where $k=1,\ldots,K_e$ and $K_e$ is the number of windows at epoch $e$. An LLM editor $f^{\mathrm{edit}}$ refines or consolidates proposals in each $\mathcal{B}_k$ into a candidate edit set $\mathcal{U}_e$, retaining the supporting cards $\mathcal{C}(u)$:
\begin{equation}
\{(u,\mathcal{C}(u)):u\in\mathcal{U}_e\}
=\bigcup_{k=1}^{K_e}f^{\mathrm{edit}}\!\left(S_e^W,\mathcal{B}_k\right).
\label{eq:candidate_edits}
\end{equation}
For each $u\in\mathcal{U}_e$, the linked cards $\mathcal{C}(u)$ determine its full set of replay trigger ranges $\mathcal{G}(u)$:
\begin{equation}
\mathcal{G}(u)=\bigcup_{C_i\in\mathcal{C}(u)}\mathcal{G}_i.
\label{eq:edit_trigger_ranges}
\end{equation}
The next phase uses $\mathcal{C}(u)$ and $\mathcal{G}(u)$ to verify $u$ in the execution contexts that motivated it.

\subsection{Replay-Guided Edit Verification}
\label{sec:edit_verification}
An evidence-grounded edit may fail to produce its intended behavioral effect. We replay candidate edits $\mathcal{U}_e$ in their supporting execution contexts and use the outcomes to verify and refine them.

\textbf{Behavioral verification.} For each $u\in\mathcal{U}_e$, we retrieve its supporting cards $\mathcal{C}(u)$ and replay the trigger ranges in $\mathcal{G}(u)$. For each $g=(x,e_g,s,t)$, we reproduce the prefix of $\tau_x^{e_g}$ preceding step $s$ to reconstruct the source execution state and interaction history. From this context, the Action Agent $\pi$ re-executes the segment under $S_e^W\oplus u$, allowing assessment of the intended correction and any newly introduced failures through comparison with the corresponding source trajectory segment.

Let $\mathcal{T}_u=\{(\tau_g^{\mathrm{src}},\tau_g^{\mathrm{rep}}):g\in\mathcal{G}(u)\}$ denote the source--replay segment pairs, where $\tau_g^{\mathrm{src}}$ is the source segment specified by $g$ and $\tau_g^{\mathrm{rep}}$ is its replayed segment. An LLM evaluator $f^{\mathrm{eval}}$ assesses these pairs using $u$ and $\mathcal{C}(u)$, returning a replay decision and associated feedback:
\begin{equation}
(r_u,h_u)=f^{\mathrm{eval}}\!\left(S_e^W,u,\mathcal{C}(u),\mathcal{T}_u\right),
\label{eq:llm_replay_assessment}
\end{equation}
where $r_u\in\{\mathrm{accept},\mathrm{reflect},\mathrm{reject}\}$ is the replay decision and $h_u$ is the corresponding feedback. The evaluator assigns \emph{accept} when replay supports the intended correction without new failures, \emph{reflect} when a localized deficiency can guide revision, and \emph{reject} otherwise. For $r_u=\mathrm{reflect}$, the editor produces a revised edit $u'=f^{\mathrm{edit}}(S_e^W,u,\mathcal{C}(u),\mathcal{T}_u,h_u)$. The revised edit retains its supporting cards, $\mathcal{C}(u')=\mathcal{C}(u)$, and undergoes another replay and evaluation.

\textbf{Verified edit consolidation.} The replay-verified edit set $\mathcal{U}_e^{+}$ contains accepted candidates and revisions accepted after reflection based on another replay and evaluation of their effects:
\begin{equation}
\mathcal{U}_e^{+}
=\{u\in\mathcal{U}_e\mid r_u=\mathrm{accept}\}
\cup
\{u'\mid u\in\mathcal{U}_e,\ r_u=\mathrm{reflect},\ r_{u'}=\mathrm{accept}\}.
\label{eq:verified_edits}
\end{equation}
Edits without local replay acceptance are excluded from $\mathcal{U}_e^{+}$, while their supporting cards remain available subject to the lifecycle updates in Section~\ref{sec:evidence_propagation}. The editor consolidates compatible edits into an ordered collection $\overline{\mathcal{U}}_e=f^{\mathrm{edit}}(S_e^W,\mathcal{U}_e^{+})$, preserving their supporting card associations. Applying $\overline{\mathcal{U}}_e$ to $S_e^W$ yields the candidate revision $\widetilde{S}_e=S_e^W\oplus\overline{\mathcal{U}}_e$ for global validation.

\subsection{Cross-Epoch Evidence Propagation}
\label{sec:evidence_propagation}
Global rejection of a candidate revision $\widetilde{S}_e$ does not establish that all its edits are ineffective on the behaviors they target. We use post-rejection replay to retain supported edits provisionally and carry unresolved cards across epochs for further synthesis and correction.

\textbf{Validation-gated carryover.} Let $\widehat{J}_{\mathcal{D}_{\mathrm{val}}}(S)$ denote the empirical performance of skill $S$ on the validation set $\mathcal{D}_{\mathrm{val}}$. Global validation compares $\widetilde{S}_e$ with the Validated Skill $S_e^V$:
\begin{equation}
S_{e+1}^{V}=
\begin{cases}
\widetilde{S}_e, &
\widehat{J}_{\mathcal{D}_{\mathrm{val}}}(\widetilde{S}_e)>
\widehat{J}_{\mathcal{D}_{\mathrm{val}}}(S_e^V),\\
S_e^V, & \text{otherwise}.
\end{cases}
\label{eq:validated_skill_update}
\end{equation}
If $\widetilde{S}_e$ is accepted, the provisional edits in $P_e$ and consolidated edits in $\overline{\mathcal{U}}_e$ are incorporated into $S_{e+1}^V$. We clear the Provisional Edit Ledger as $P_{e+1}=\varnothing$ and set the next Tracked Edits to $\mathcal{Q}_{e+1}=P_e\cup\overline{\mathcal{U}}_e$, preserving their supporting cards for cross-epoch comparison of training trajectories.

If $\widetilde{S}_e$ is rejected, $S_{e+1}^V=S_e^V$. Each consolidated edit $v\in\overline{\mathcal{U}}_e$ undergoes post-rejection replay using its supporting cards $\mathcal{C}(v)$ and trigger ranges $\mathcal{G}(v)$. We reconstruct the source contexts as in Section~\ref{sec:edit_verification}, but replay under $S_e^V\oplus v$ to assess whether $v$ remains effective without relying on the provisional edits in $P_e$. The evaluator $f^{\mathrm{eval}}$ returns a decision $r_v^{\mathrm{post}}$. Edits initially assigned \emph{reflect} are revised by $f^{\mathrm{edit}}$ with their evidence links preserved, replayed under $S_e^V\oplus v'$, and reassessed by $f^{\mathrm{eval}}$ to obtain a new decision on $v'$. The retained edit set $\mathcal{U}_e^{\mathrm{post},+}$ is:
\begin{equation}
\mathcal{U}_e^{\mathrm{post},+}
=\{v\in\overline{\mathcal{U}}_e\mid r_v^{\mathrm{post}}=\mathrm{accept}\}
\cup
\{v'\mid v\in\overline{\mathcal{U}}_e,\ 
r_v^{\mathrm{post}}=\mathrm{reflect},\ 
r_{v'}^{\mathrm{post}}=\mathrm{accept}\}.
\label{eq:post_rejection_retained_edits}
\end{equation}
Edits in $\mathcal{U}_e^{\mathrm{post},+}$ replace the entries they target in $P_e$, thereby updating retained edits with evidence-grounded corrections, while the remaining retained edits are appended in order, yielding $P_{e+1}$. Edits without acceptance are not added to the ledger. We set $\mathcal{Q}_{e+1}=P_{e+1}$ and form the Working Skill $S_{e+1}^W=S_{e+1}^V\oplus P_{e+1}$, enabling continued training and refinement while $S_{e+1}^V$ remains unchanged.

\textbf{Evidence reuse and correction.} Validation and replay outcomes govern card availability in the Evidence Pool $\mathcal{C}_e$. \emph{Active} cards remain eligible for edit synthesis. Cards supporting provisional edits in $P_{e+1}$ are \emph{protected}: they are excluded from Evidence Windows $\mathcal{B}_k$, but their task and trajectory references remain available for cross-epoch comparison. If $\widetilde{S}_e$ is accepted, cards supporting the incorporated edits are \emph{archived}, likewise excluded from new windows but retained for comparison. Repeated replay rejection can mark active cards as \emph{stale}, excluding them from new synthesis.

These lifecycle updates determine the carried evidence cards $\mathcal{C}_{e+1}^{\mathrm{carry}}$. In the next epoch, the Action Agent $\pi$ uses $S_{e+1}^W$ to collect $\mathcal{R}_{e+1}$. Following Section~\ref{sec:edit_construction}, $f^{\mathrm{ext}}$ extracts $\mathcal{C}_{e+1}^{\mathrm{traj}}$ from these trajectories and $\mathcal{C}_{e+1}^{\mathrm{con}}$ through comparisons guided by $\mathcal{Q}_{e+1}$ and its supporting cards. Both sets join $\mathcal{C}_{e+1}^{\mathrm{carry}}$ to form $\mathcal{C}_{e+1}$ via Eq.~\ref{eq:evidence_propagation}, supporting further edit synthesis and targeted correction.

After $E$ evolution epochs, the latest Validated Skill $S_{E+1}^V$ is frozen for inference:
\begin{equation}
S^\star=S_{E+1}^V,\qquad
\tau_x\sim\pi(\cdot\mid x,S^\star),\quad
x\in\mathcal{D}_{\mathrm{test}}.
\label{eq:final_skill_inference}
\end{equation}
Unpromoted provisional edits are excluded from $S^\star$, and do not trigger further skill evolution.

\vspace{-5pt}
\section{Experiments}
\label{sec:experiments}
\vspace{-5pt}
In this section, we conduct experiments to evaluate the effectiveness of \method{} and examine how its evidence-grounded mechanisms support continual skill evolution. Specifically, we aim to answer the following questions. \textbf{Q1 (Overall Performance):} How does \method{} perform compared to existing skill-evolution methods across different benchmarks and LLM backbones? \textbf{Q2 (Ablation Study):} How do trigger-range replay and cross-epoch evidence propagation contribute to the overall performance? \textbf{Q3 (Mechanism Analysis):} How does replay filter and refine proposed edits, and how do retained evidence and supported edits contribute to subsequent evolution?

\vspace{-5pt}
\subsection{Experimental Setting}
\label{sec:experimental_setting}

We summarize the datasets, evaluation protocol, comparison methods, and implementation settings used across the three interactive benchmarks below, with full details in Appendix~\ref{app:experimental_details}.

\paragraph{Datasets and Evaluation.}
We evaluate \method{} on three interactive benchmarks: AppWorld~\citep{appworld} for task execution through application APIs, ScienceWorld~\citep{scienceworld} for scientific experimentation, and ALFWorld~\citep{alfworld} for household task execution in text-based environments. We use disjoint training, validation, and test splits for skill evolution, revision selection, and final evaluation, respectively. Performance is measured by task-success accuracy (Acc., \%), the percentage of test tasks satisfying each benchmark's full success criterion. All methods are evaluated on the same fixed test tasks without further skill evolution.

\paragraph{Baselines and Implementation.}
We compare against two groups: (i) \emph{Non-Evolving Baselines}, including NoSkill, which uses no external skills, and LLM Skill, which uses a fixed initial LLM-generated skill; and (ii) \emph{Skill-Evolution Methods}, including Trace2Skill~\citep{trace2skill2026}, EvoSkill~\citep{evoskill2026}, SkillGrad~\citep{skillgrad2026}, and SkillOpt~\citep{skillopt2026}. We use six Action-Agent backbones: GPT-5.5, GPT-5.4, GPT-5.4-mini, Qwen3.6-35B-A3B, Qwen3.5-9B, and Qwen3.5-4B, with the Action Agent backbone, task splits, environment configuration, and evaluation protocol held fixed across methods within each benchmark and backbone setting.

\subsection{Main Results}
\label{sec:overall_results}

\providecommand{\method}{\textsc{EviSkill}}
\providecommand{\bestresult}[1]{\textbf{#1}}
\providecommand{\secondresult}[1]{\underline{#1}}
\definecolor{eviskillblue}{RGB}{232,242,255}

\begin{table}[t]
\caption{Task-success accuracy (\%) across three interactive benchmarks. $\Delta_{\mathrm{NS}}$ denotes the percentage-point change relative to No skill for the same model and dataset. \textbf{Bold} and \underline{underlined} values indicate the best and second-highest distinct accuracies within each model--dataset setting.}
\vspace{-10pt}
\label{tab:main-results}
\centering
\setlength{\tabcolsep}{1.0pt}
\renewcommand{\arraystretch}{1.02}
\begin{tabularx}{\columnwidth}{@{}
>{\hsize=1.35\hsize\raggedright\arraybackslash}X
>{\hsize=.95\hsize\raggedright\arraybackslash}X
*{6}{>{\hsize=.95\hsize\raggedleft\arraybackslash}X}@{}}
\toprule
\multicolumn{1}{c}{Model} &
\multicolumn{1}{c}{Method} &
\multicolumn{2}{c}{ALFWorld (134)} &
\multicolumn{2}{c}{AppWorld (168)} &
\multicolumn{2}{c}{ScienceWorld (211)} \\
\cmidrule(lr){3-4}\cmidrule(lr){5-6}\cmidrule(lr){7-8}
& & Acc.$\uparrow$ & $\Delta_{\mathrm{NS}}$ &
Acc.$\uparrow$ & $\Delta_{\mathrm{NS}}$ &
Acc.$\uparrow$ & $\Delta_{\mathrm{NS}}$ \\
\midrule
\multirow{7}{*}{\centering GPT-5.5}
& No Skill & 90.30 & 0.00 & 85.71 & 0.00 & 76.30 & 0.00 \\
& LLM Skill & 94.03 & +3.73 & \secondresult{94.05} & +8.33 & 76.78 & +0.47 \\
& Trace2Skill & 94.03 & +3.73 & 86.31 & +0.60 & 65.40 & -10.90 \\
& EvoSkill & 94.78 & +4.48 & 91.07 & +5.36 & 54.50 & -21.80 \\
& SkillGrad & \secondresult{95.52} & +5.22 & 86.31 & +0.60 & 67.30 & -9.00 \\
& SkillOpt & 94.78 & +4.48 & 92.86 & +7.14 & \secondresult{80.09} & +3.79 \\
\rowcolor{eviskillblue}
& \method{} & \bestresult{96.27} & +5.97 & \bestresult{94.64} & +8.93 & \bestresult{83.41} & +7.11 \\
\midrule
\addlinespace[1.0pt]
\multirow{7}{*}{\centering GPT-5.4}
& No Skill & 75.37 & 0.00 & 88.10 & 0.00 & 52.13 & 0.00 \\
& LLM Skill & 89.55 & +14.18 & 88.69 & +0.60 & \secondresult{66.35} & +14.22 \\
& Trace2Skill & \secondresult{94.78} & +19.40 & 85.71 & -2.38 & 56.40 & +4.27 \\
& EvoSkill & 90.30 & +14.93 & 79.17 & -8.93 & 63.03 & +10.90 \\
& SkillGrad & 94.03 & +18.66 & 83.93 & -4.17 & 53.55 & +1.42 \\
& SkillOpt & \secondresult{94.78} & +19.40 & \secondresult{89.29} & +1.19 & 63.03 & +10.90 \\
\rowcolor{eviskillblue}
& \method{} & \bestresult{95.52} & +20.15 & \bestresult{90.48} & +2.38 & \bestresult{76.78} & +24.64 \\
\midrule
\addlinespace[1.0pt]
\multirow{7}{*}{\centering Qwen3.5-9B}
& No Skill & 54.48 & 0.00 & 25.60 & 0.00 & 8.53 & 0.00 \\
& LLM Skill & 76.12 & +21.64 & 43.45 & +17.86 & 9.00 & +0.47 \\
& Trace2Skill & 75.37 & +20.90 & 45.24 & +19.64 & 19.91 & +11.37 \\
& EvoSkill & 76.87 & +22.39 & 32.74 & +7.14 & 17.06 & +8.53 \\
& SkillGrad & \secondresult{87.31} & +32.84 & \secondresult{46.43} & +20.83 & 14.22 & +5.69 \\
& SkillOpt & 86.57 & +32.09 & \bestresult{47.62} & +22.02 & \secondresult{22.27} & +13.74 \\
\rowcolor{eviskillblue}
& \method{} & \bestresult{91.04} & +36.57 & 43.45 & +17.86 & \bestresult{37.91} & +29.38 \\
\midrule
\addlinespace[1.0pt]
\multirow[c]{7}{=}{Qwen3.5-4B}
& No Skill & 43.28 & 0.00 & 18.45 & 0.00 & 4.27 & 0.00 \\
& LLM Skill & 75.37 & +32.09 & 28.57 & +10.12 & 3.79 & -0.47 \\
& Trace2Skill & 82.84 & +39.55 & \bestresult{32.14} & +13.69 & 9.48 & +5.21 \\
& EvoSkill & 73.88 & +30.60 & 27.98 & +9.52 & 2.84 & -1.42 \\
& SkillGrad & \secondresult{84.33} & +41.04 & 30.36 & +11.90 & 6.64 & +2.37 \\
& SkillOpt & 83.58 & +40.30 & \secondresult{30.95} & +12.50 & \secondresult{14.69} & +10.43 \\
\rowcolor{eviskillblue}
& \method{} & \bestresult{85.07} & +41.79 & \bestresult{32.14} & +13.69 & \bestresult{20.85} & +16.59 \\
\bottomrule
\end{tabularx}
\vspace{-10pt}
\end{table}
To answer Q1, we compare \method{} with reference settings and skill-evolution baselines across three interactive benchmarks and four LLM backbones. Table~\ref{tab:main-results} presents these results. The complete comparison across six backbones is reported in Table~\ref{tab:complete-main-results} in Appendix~\ref{app:additional_results}.

\noindent\textbf{Obs. 1. Grounding skill updates in execution evidence yields broad performance gains.}
\method{} achieves the best accuracy in 14 of the 18 model--dataset settings and ranks among the top two methods in 16. It improves over NoSkill in all settings, with an average gain of 17.93 percentage points. On ScienceWorld, it exceeds the strongest baseline by 15.64 points with Qwen3.5-9B and 10.43 points with GPT-5.4. A central challenge in skill evolution is the gap between \textbf{plausible guidance and effective execution}: a revision inferred from past trajectories may omit necessary conditions or fail to change the behavior it targets. The gains are consistent with the value of assessing updates through their \textbf{actual behavioral effects}, which helps distinguish executable improvements from seemingly reasonable suggestions. This benefit extends across both model families, although the advantage over competing methods is less consistent on AppWorld.

\noindent\textbf{Obs. 2. Evolving a skill does not guarantee improvement over its initial version.}
Several evolution baselines underperform LLM Skill. On ScienceWorld with GPT-5.5, for example, the initial skill achieves 76.78\%, whereas Trace2Skill, EvoSkill, and SkillGrad obtain 65.40\%, 54.50\%, and 67.30\%, respectively. In contrast, \method{} reaches 83.41\% and improves over LLM Skill in 17 settings while matching it in the remaining one. These results highlight the difficulty of \textbf{preserving useful knowledge} while incorporating new experience. Individual changes can have different effects, so rejecting a complete revision may discard useful corrections, while accepting it may introduce ineffective guidance. Retaining supported changes for \textbf{continued refinement across epochs} offers a way to preserve progress while reconsidering uncertain updates. The consistent gains of \method{} support this explanation, which we examine further below.

\subsection{Ablation Study}
\label{sec:ablation}

\begin{wrapfigure}{l}{0.44\textwidth}
    \centering
    \vspace{-15pt}
    \includegraphics[width=\linewidth]{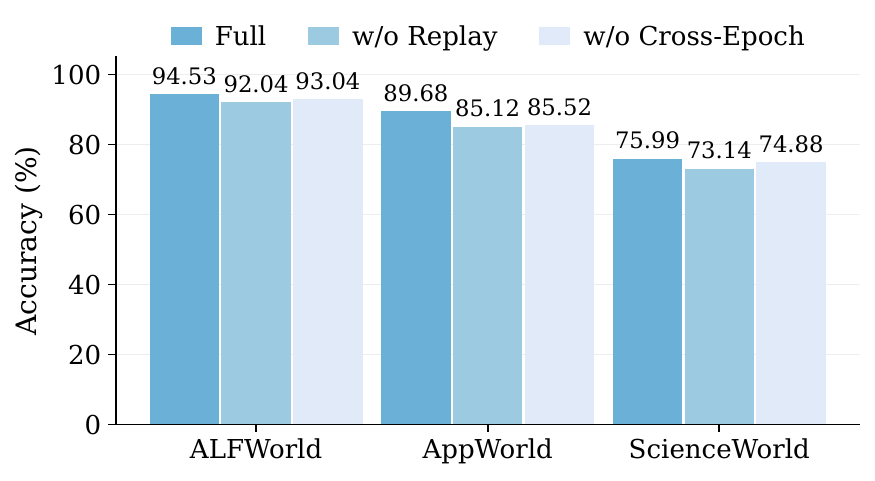}
     \vspace{-15pt}
    \caption{Component ablations averaged over the three GPT backbones.}
    \label{fig:ablation}
    \vspace{-15pt}
\end{wrapfigure}

To answer Q2, we examine two components of \method{}: (i) \emph{w/o Replay}, which removes Replay-Guided Edit Verification, and (ii) \emph{w/o Cross-Epoch}, which disables Cross-Epoch Evidence Propagation. Figure~\ref{fig:ablation} reports task-success accuracy averaged over the three GPT backbones. Complete results across all six backbones are reported in Figure~\ref{fig:complete_ablation_results} in Appendix~\ref{app:additional_results}.

\noindent\textbf{Obs. 3. Replay-guided verification helps turn proposed edits into effective behavioral corrections.}
Removing replay reduces average accuracy by 2.49, 4.56, and 2.85 percentage points on ALFWorld, AppWorld, and ScienceWorld, respectively. The consistent decreases support the importance of \textbf{verifying edits through execution}. An edit may appear consistent with its source trajectory yet omit a necessary condition or fail to correct the targeted behavior. Re-executing the associated trajectory ranges exposes such discrepancies, providing feedback for filtering unsupported edits and refining correctable proposals. The results indicate that assessing an edit's observed effects contributes beyond constructing it from past experience.

\medskip
\noindent\textbf{Obs. 4. Cross-epoch propagation preserves useful progress beyond individual revision attempts.}
Disabling cross-epoch propagation reduces accuracy on all three benchmarks, with the largest decrease on AppWorld, from 89.68\% to 85.52\%. This supports the value of \textbf{retaining and revisiting evolutionary information}: rejection of a complete revision does not establish that all its constituent edits are ineffective, nor that unresolved evidence cannot support later improvements. Carrying supported edits and unresolved evidence forward allows subsequent experience to reinforce, refine, or invalidate earlier proposals. Together, the two ablations show that checking current edits and preserving information for future refinement both contribute to effective skill evolution.


\begin{figure}[h]
    \centering
    \begin{subfigure}[b]{0.48\linewidth}
        \centering
        \includegraphics[width=\linewidth]{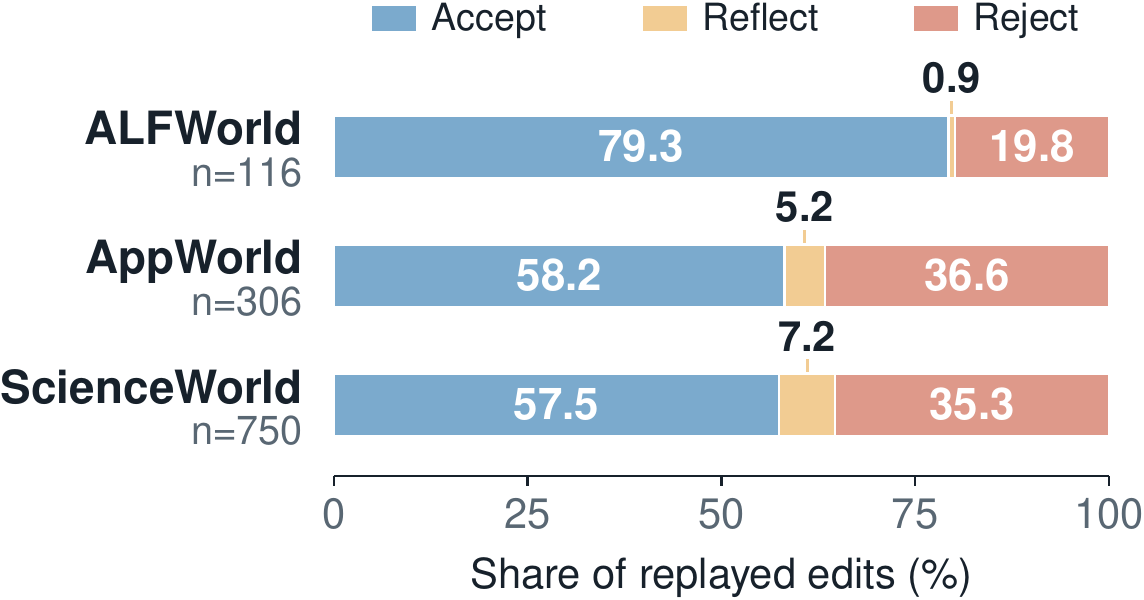}
        \caption{Initial decisions over candidate edits.}
        \label{fig:replay_edit_outcomes_a}
    \end{subfigure}
    \hfill
    \begin{subfigure}[b]{0.48\linewidth}
        \centering
        \includegraphics[width=\linewidth]{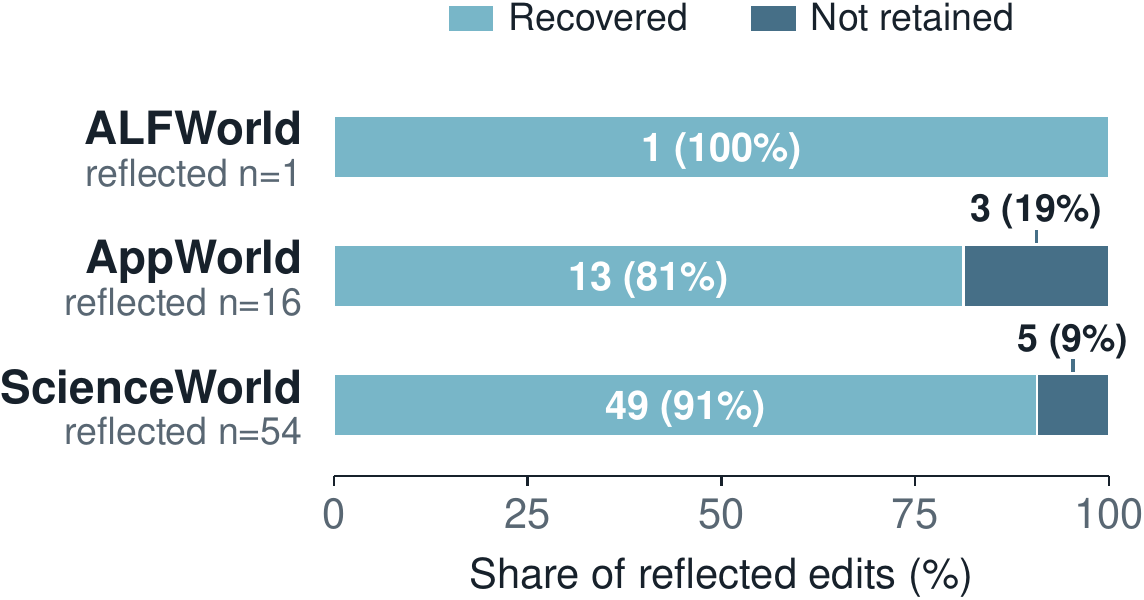}
        \caption{Outcomes after reflection assignment.}
        \label{fig:replay_edit_outcomes_b}
    \end{subfigure}

    \caption{Replay-guided edit verification. (a) Initial decisions over all candidate edits. (b) Refinement outcomes among edits initially assigned to reflection, showing counts and within-group percentages. Acceptance indicates local replay support, not global revision acceptance.}
    \label{fig:replay_edit_outcomes}
    \vspace{-15pt}
\end{figure}

\subsection{Mechanism Analysis}
\label{sec:mechanism_analysis}

To answer Q3, we analyze 18 runs spanning 72 epochs, focusing on replay-based edit verification and refinement, and the contribution of retained edits and evidence to subsequent evolution.

\subsubsection{Replay-Guided Edit Verification}
\label{sec:evidence_grounded_mechanism_analysis}

Figure~\ref{fig:replay_edit_outcomes} summarizes initial replay decisions and subsequent refinement outcomes for edits assigned \emph{reflect}, pooling edit counts across runs and epochs within each benchmark.

\noindent\textbf{Obs. 5. Replay identifies unsupported proposals while enabling feedback-guided correction.}
As shown in Figure~\ref{fig:replay_edit_outcomes_a}, although candidate edits are grounded in training trajectories, a substantial fraction do not pass their initial replay: 20.7\% on ALFWorld, 41.8\% on AppWorld, and 42.5\% on ScienceWorld are assigned to reflection or rejection. These outcomes reveal a gap between \textbf{trajectory-derived guidance and demonstrated behavioral improvement}, supporting the need to verify edits before incorporating them into a candidate revision. However, an initially unsupported edit is not necessarily beyond correction. Figure~\ref{fig:replay_edit_outcomes_b} shows that feedback-guided refinement recovers 13 of 16 reflected edits on AppWorld and 49 of 54 on ScienceWorld, as well as the single reflected edit on ALFWorld. Replay thus provides both a basis for \textbf{filtering unsupported edits} and actionable feedback for \textbf{refining correctable proposals}. These recovery rates describe the subset selected for reflection after the initial replay, rather than all initially unsupported edits.

\subsubsection{Cross-Epoch Evidence Propagation}
\label{sec:cross_epoch_mechanism_analysis}

\begin{wrapfigure}{l}{0.42\linewidth}
    \vspace{-\intextsep}
    \centering
    \includegraphics[width=\linewidth]{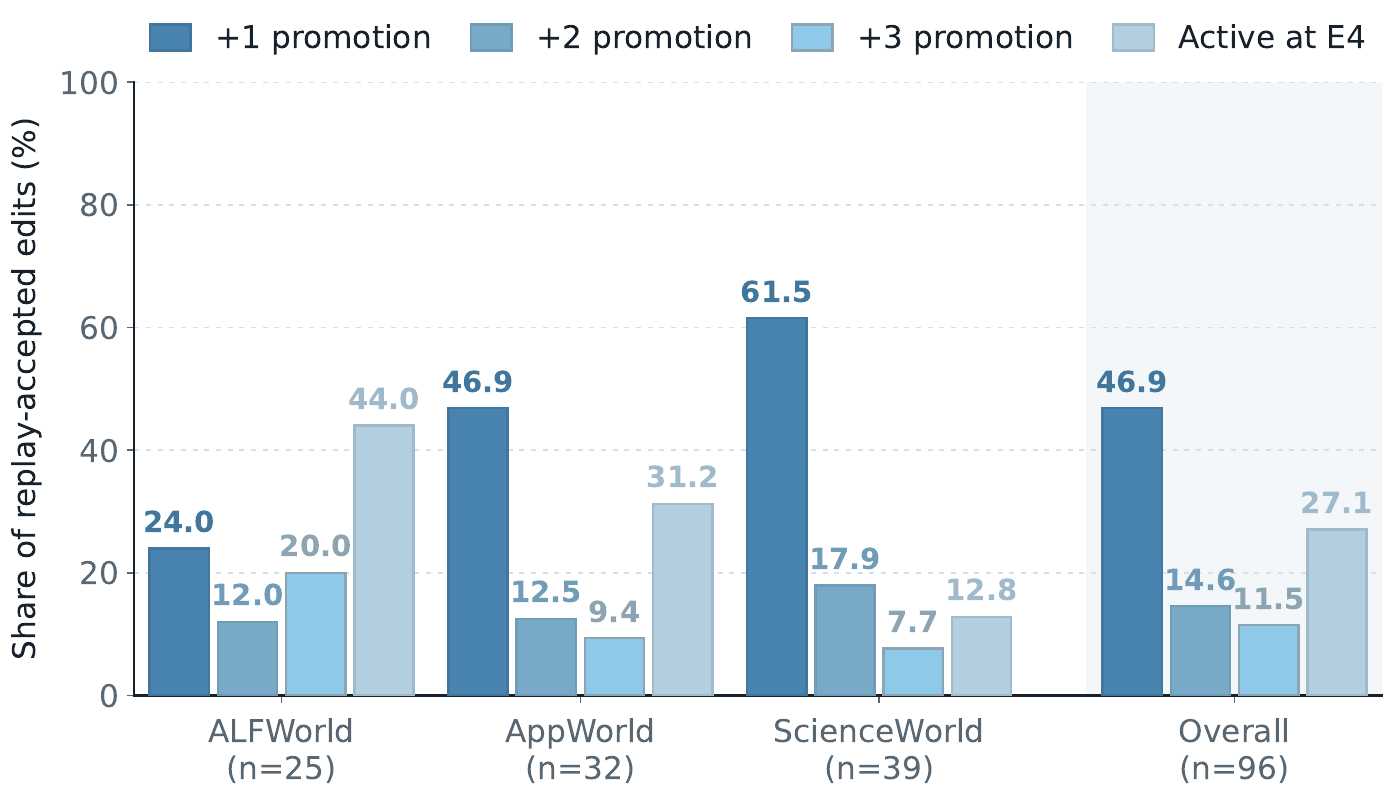}
    \vspace{-10pt}
    \caption{Promotion of the 96 provisionally retained edits.}
    \label{fig:cross_epoch_promotion}
    \vspace{-10pt}
\end{wrapfigure}

\begin{wrapfigure}{r}{0.42\linewidth}
    \vspace{-\intextsep}
    \centering
    \includegraphics[width=\linewidth]{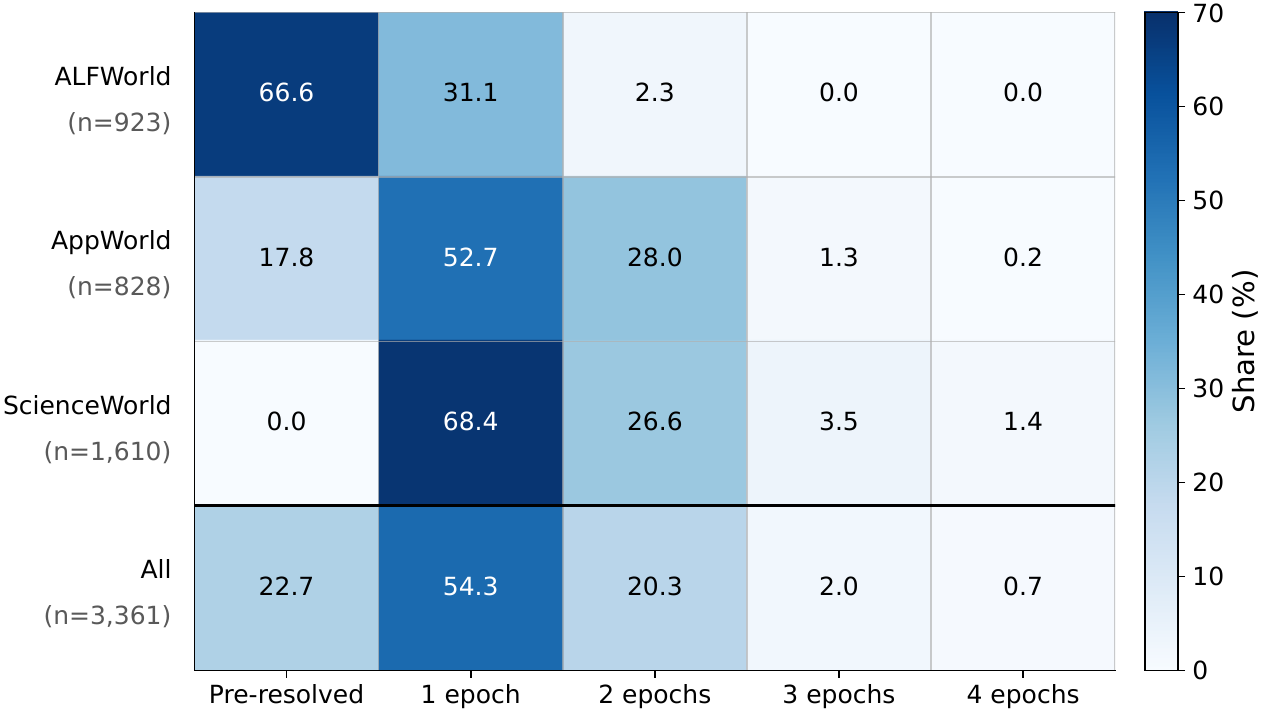}
    \caption{Number of distinct epochs in which each Evidence Card is selected for an Evidence Window.}
    \label{fig:cross_epoch_evidence_reuse}
\end{wrapfigure}

We examine whether replay-accepted edits retained after global rejection are later incorporated into a Validated Skill and whether Evidence Cards continue to support edit construction across epochs. Figures~\ref{fig:cross_epoch_promotion} and~\ref{fig:cross_epoch_evidence_reuse} summarize these two outcomes across three benchmarks and four epochs.

\noindent\textbf{Obs. 6. Globally rejected revisions contain edits that contribute to later accepted skills.}
\par
\noindent
We analyze the 96 edits provisionally retained after post-rejection replay. As shown in Figure~\ref{fig:cross_epoch_promotion}, 46.9\% enter the Validated Skill one epoch later, 14.6\% do so two epochs later, and 11.5\% do so three epochs later. Thus, \textbf{global revision rejection does not invalidate every constituent edit}: 72.9\% of these locally supported edits are subsequently incorporated into globally accepted skills. Preserving them provisionally allows useful corrections to survive an unsuccessful revision attempt and contribute to later evolution. Another 27.1\% remain provisional when the four-epoch budget ends, so their eventual promotion remains unobserved.

\noindent\textbf{Obs. 7. Evidence continues to support edit construction beyond a single epoch.}
\par
\noindent
Across all benchmarks, 23.0\% of unique Evidence Cards are selected for Evidence Windows in at least two distinct epochs, while 54.3\% are selected in exactly one (Figure~\ref{fig:cross_epoch_evidence_reuse}). This demonstrates \textbf{continued use of source evidence} across evolution epochs, allowing later edit construction to revisit previously collected experience. The extent of reuse varies across environments: 66.6\% of ALFWorld Cards are pre-resolved, whereas AppWorld and ScienceWorld contain larger shares selected across multiple epochs. These results and the subsequent promotion of retained edits show that cross-epoch propagation preserves both proposed corrections and the evidence available for further refinement.



\section{Conclusion}

We introduced \method{}, an evidence-grounded framework for continual skill evolution. \method{} links skill edits to their supporting execution contexts and verifies their intended effects through replay of the relevant trigger ranges. When global validation rejects a revision, post-rejection replay selectively retains supported edits for further refinement. Unresolved evidence and cross-epoch trajectory comparisons guide subsequent synthesis and correction. Experiments on three interactive benchmarks across LLM backbones demonstrate consistent gains over agents without skills and competitive performance against skill-evolution baselines. These results support verifying edits through execution and reassessing their value beyond a single revision outcome.

\subsection*{AI Use Statement}

We used generative AI tools to assist with manuscript organization and language refinement. We independently formulated the hypotheses, designed and implemented the method and experiments, processed the datasets, and interpreted the results without generative AI assistance. We checked the AI-assisted writing for accuracy and consistency with the underlying research and take full responsibility for the paper’s text, claims, and artifacts.

\subsection*{Ethics Statement}
The three benchmarks used in our experiments, AppWorld, ScienceWorld, and ALFWorld, are publicly available and widely used. Our research adheres to the ICLR Code of Ethics, particularly regarding data privacy, transparent reporting, and research integrity.

\subsection*{Reproducibility Statement}
The method section describes the evidence-grounded skill evolution framework. Appendix~\ref{app:experimental_details} provides dataset splits and baseline descriptions. Appendix~\ref{app:method_details} details the algorithm, evidence and edit representations, replay implementation, lifecycle management, model configurations, and evaluation settings. Appendix~\ref{app:preliminary_details} describes the procedures for the two preliminary studies, while Appendix~\ref{app:prompts} provides the core prompts. Our code is available at \url{https://github.com/Zhouyaner/eviskill}.

\bibliographystyle{iclr2027_conference}
\bibliography{iclr2027_conference}

\newpage

\appendix
\clearpage
\etocdepthtag.toc{appendix}

\begingroup
\renewcommand{\contentsname}{Appendix Contents}
\hypersetup{hidelinks}

\etocsettagdepth{main}{none}
\etocsettagdepth{appendix}{subsection}
\tableofcontents

\endgroup
\clearpage

\section{Related Work}

\paragraph{Agent Skills.} Agent skills provide reusable procedural guidance that an LLM agent can load when performing a task, without modifying the model parameters~\citep{voyager,skillnet2026,mementoskills,agentskillssurvey}. They can specify tool-use procedures, action sequences, task constraints, and recovery steps, and may include supporting scripts or resources~\citep{skillcraft,skillreducer}. Such guidance has been used in embodied interaction, web navigation, and coding workflows~\citep{voyager,skillweaver,codeskill2026,skillfoundry}. Studies across diverse tasks further show that the benefit of a skill depends on whether its guidance helps the agent execute the task effectively~\citep{skillsbench,skillflow,sweskillsbench,skilllearnbench,wildskills}. These settings motivate treating skills as reusable artifacts whose usefulness must be assessed through agent execution in the intended task environments.

\paragraph{Experience Memory.} Agents can also reuse past interactions through verbal reflections, extracted lessons, and stored workflows~\citep{reflexion,expel,AWM,generativeagents,mementoskills,memorybank,synapse}. Procedural and reasoning memory methods further organize experience into guidance that can be retrieved or updated during later tasks~\citep{memp,remember,reasoningbank,memskill,procmem}. Related approaches evolve the agent's external context or manage experience across an interaction lifecycle from collection to reuse~\citep{agentic,evolver2025,memgpt,dynamiccheatsheet}. This line of work informs how execution history can support future decisions.

\paragraph{Skill Evolution.} Existing methods use execution experience to revise skills in different ways~\citep{skillaudit2026,skillcat2026,skillprox2026,mementoskills,skillclaw}. Trace2Skill extracts trajectory-local patches and consolidates them into a reusable artifact~\citep{trace2skill2026}; SkillAdaptor and SkillRevise use execution failures and trace-conditioned feedback to guide revisions~\citep{skilladaptor2026,skillrevise2026,skillforge}. SkillGrad converts diagnoses into textual update signals and accumulates recurring patterns across iterations~\citep{skillgrad2026}. EvoSkill selects skill candidates through iterative evaluation, while SkillOpt applies bounded edits, uses held-out validation, and retains rejected edits as feedback for later updates~\citep{evoskill2026,skillopt2026}.

\section{Experimental Details}
\label{app:experimental_details}
\subsection{Datasets and Task Splits}
\label{app:datasets}

\begin{wraptable}{l}{0.55\textwidth}
    \centering
    \small
    \caption{Task families and dataset splits.}
    \label{tab:dataset-splits}
    \begin{tabular*}{\linewidth}{@{\extracolsep{\fill}}lcrrr@{}}
        \toprule
        Dataset & \shortstack{Task Families} & Train & Val. & Test \\
        \midrule
        AppWorld     & -  & 90 & 57 & 168 \\
        ScienceWorld & 24 & 96 & 48 & 211 \\
        ALFWorld     & 6  & 96 & 48 & 134 \\
        \bottomrule
    \end{tabular*}
\end{wraptable}
We evaluate on three interactive benchmarks: AppWorld~\citep{appworld}, ScienceWorld~\citep{scienceworld}, and ALFWorld~\citep{alfworld}, covering application API use, scientific experimentation, and household task execution, respectively. These benchmarks evaluate whether evolved skills provide reusable procedural guidance for multi-step interactions across different environments with distinct actions and observations. Table~\ref{tab:dataset-splits} summarizes the task splits. We use disjoint training, validation, and test splits for trajectory collection and skill evolution, global validation and revision selection, and final evaluation, respectively.
\begin{itemize}
    \item \textbf{AppWorld.}
    AppWorld is a controllable environment for evaluating interactive agents through realistic application APIs. Each task requires multi-step operations across simulated applications, where success depends on planning API calls, maintaining intermediate states, and satisfying the final task objective. We directly use the complete official \texttt{train}, \texttt{dev}, and \texttt{test\_normal} splits, without additional sampling. These splits contain 90, 57, and 168 tasks, respectively, and have no overlapping scenario families.

    \item \textbf{ScienceWorld.}
    ScienceWorld evaluates agents on interactive scientific experimentation tasks. Agents must manipulate objects, conduct experiments, interpret environmental feedback, and complete task-specific goals. These tasks emphasize procedural reasoning and recovery from intermediate failures. We sample four instances per task family from the official \texttt{train} split and two per family from \texttt{dev}, yielding 96 training and 48 validation tasks across all 24 families. We use the complete official \texttt{test} split of 211 tasks.

    \item \textbf{ALFWorld.}
    ALFWorld is a text-based embodied environment for household task completion. Agents navigate environments and interact with objects through textual actions to accomplish goals such as locating, manipulating, and placing items. Its recurring task procedures provide a setting for evaluating reusable execution strategies. For each of the six task families, we select 16 distinct scenarios from the official \texttt{train} split and eight from \texttt{valid\_seen}, yielding 96 training and 48 validation tasks. We retain only one trial per scenario within each sampled split. The complete official \texttt{valid\_unseen} split provides 134 test tasks. We check the manifests to ensure no gamefile appears in more than one split.
\end{itemize}

\subsection{Baselines}
\label{app:baselines}

We group the evaluated methods into \textbf{non-evolving baselines} and \textbf{skill-evolution methods}. The former use either no external skill or a fixed LLM-generated skill, while the latter update skills using execution experience. Table~\ref{tab:baseline-overview} summarizes their core mechanisms.

\begin{table}[h]
\centering
\small
\caption{Overview of the evaluated methods and their skill-evolution strategies.}
\label{tab:baseline-overview}
\renewcommand{\arraystretch}{1.18}
\setlength{\tabcolsep}{5pt}
\begin{tabular*}{\textwidth}{@{\extracolsep{\fill}}lll@{}}
\toprule
Method & Evolution strategy & Core mechanism \\
\midrule
\multicolumn{3}{@{}l}{\emph{Non-Evolving Baselines}} \\
NoSkill   & None & No external skill \\
LLM Skill & None & Fixed LLM-generated guidance \\
\midrule
\multicolumn{3}{@{}l}{\emph{Skill-Evolution Methods}} \\
Trace2Skill & Trajectory distillation & Lesson extraction and consolidation \\
EvoSkill    & Iterative discovery & Skill generation and refinement \\
SkillGrad   & Textual-gradient optimization & Update accumulation and patching \\
SkillOpt    & Validation-guided revision & Candidate evaluation and selection \\
\method{} (Ours) & Evidence-grounded evolution & Replay verification and cross-epoch reuse \\
\bottomrule
\end{tabular*}
\end{table}

\paragraph{Non-Evolving Baselines.}
\begin{itemize}
    \item \textbf{NoSkill.}
    The Action Agent executes each task without an external skill document or skill-related memory. Interaction trajectories are not reused across tasks, and no skill is generated or updated. This setting measures the underlying agent's task-solving performance across the three benchmarks before introducing reusable procedural guidance.

    \item \textbf{LLM Skill.}
    The Action Agent receives the initial LLM-generated skill $S_0$, which remains fixed throughout evaluation. The skill provides procedural guidance but incorporates no feedback from subsequent interactions. This setting measures the benefit of skill initialization and provides the reference for assessing gains from further evolution.
\end{itemize}

\paragraph{Skill-Evolution Methods.}
\begin{itemize}
    \item \textbf{Trace2Skill}~\citep{trace2skill2026}.
    Trace2Skill derives reusable guidance from local execution experience. It identifies procedural lessons in agent trajectories, translates them into skill patches, and consolidates them into transferable skill artifacts. Its emphasis is on converting concrete interaction experience into instructions that can guide subsequent tasks.

    \item \textbf{EvoSkill}~\citep{evoskill2026}.
    EvoSkill iteratively discovers and refines reusable skills from accumulated interactions. Task outcomes and failure analyses guide candidate improvements, which undergo evaluation and consolidation as the skill set evolves. Its emphasis is on the repeated discovery and integration of useful procedures across tasks.

    \item \textbf{SkillGrad}~\citep{skillgrad2026}.
    SkillGrad represents execution diagnoses as textual gradient-like update signals. It accumulates recurring improvement directions across iterations and uses them to construct structured patches to the skill. This mechanism connects observed execution weaknesses to targeted changes in procedural guidance.

    \item \textbf{SkillOpt}~\citep{skillopt2026}.
    SkillOpt treats skill refinement as candidate revision and selection. It uses execution feedback to modify an existing skill, evaluates the resulting candidates, and retains revisions that improve validation performance. Its central mechanism is performance-based selection at the skill-revision level.

    \item \textbf{\method{} (Ours).}
    \method{} associates proposed edits with Replayable Evidence Cards that preserve their supporting task contexts and trajectory ranges for reconstructing source execution states during replay. Targeted replay examines the edits' behavioral effects and provides feedback for refinement. Across epochs, unresolved evidence and locally supported edits remain available for continued evolution, while global validation determines whether a candidate revision is incorporated into the Validated Skill. This separates local support for individual edits from acceptance of the complete skill revision.
\end{itemize}

\section{Method Implementation Details}
\label{app:method_details}
\subsection{Algorithm}
\label{app:overall_algorithm}

Algorithm~\ref{alg:eviskill_overall} summarizes the evolution procedure of \method{}. Across epochs, it maintains the Validated Skill $S_e^V$, Provisional Edit Ledger $P_e$, carried cards $\mathcal{C}_e^{\mathrm{carry}}$, and Tracked Edits $\mathcal{Q}_e$. Each edit $u$ retains its supporting cards $\mathcal{C}(u)$, which identify execution contexts for replay and tasks for cross-epoch comparison using complete trajectory pairs from adjacent epochs. Evidence extraction, edit synthesis, and replay assessment use $f^{\mathrm{ext}}$, $f^{\mathrm{edit}}$, and $f^{\mathrm{eval}}$, respectively.

\begin{algorithm*}[t]
\caption{Overall procedure of \method{}}
\label{alg:eviskill_overall}
\small
\begin{algorithmic}[1]
\Require $\pi$, $f^{\mathrm{ext}}$, $f^{\mathrm{edit}}$, $f^{\mathrm{eval}}$, $\mathcal{D}_{\mathrm{tr}}$, $\mathcal{D}_{\mathrm{val}}$, $S_0$, and $E$
\Ensure Final Validated Skill $S^\star$
\State $S_1^V\gets S_0$, $P_1\gets\varnothing$, $\mathcal{Q}_1\gets\varnothing$, $\mathcal{C}_1^{\mathrm{carry}}\gets\varnothing$
\For{$e=1,\ldots,E$}
    \State $S_e^W\gets S_e^V\oplus P_e$
    \State $\mathcal{R}_e\gets\{\tau_x^e\sim\pi(\cdot\mid x,S_e^W)\mid x\in\mathcal{D}_{\mathrm{tr}}\}$
    \State $\mathcal{C}_e^{\mathrm{traj}}\gets f^{\mathrm{ext}}(S_e^W,\mathcal{R}_e)$
    \If{$e>1\land\mathcal{Q}_e\neq\varnothing$}
        \State $\mathcal{X}_e\gets$ training tasks referenced by $\bigcup_{u\in\mathcal{Q}_e}\mathcal{C}(u)$
        \State $\mathcal{R}_e^{\mathrm{cmp}}\gets\{(\tau_x^{e-1},\tau_x^e)\mid x\in\mathcal{X}_e\}$
        \State $\mathcal{C}_e^{\mathrm{con}}\gets f^{\mathrm{ext}}\!\left(S_e^W,\{(u,\mathcal{C}(u)):u\in\mathcal{Q}_e\},\mathcal{R}_e^{\mathrm{cmp}}\right)$
    \Else
        \State $\mathcal{C}_e^{\mathrm{con}}\gets\varnothing$
    \EndIf
    \State $\mathcal{C}_e\gets\mathcal{C}_e^{\mathrm{carry}}\cup\mathcal{C}_e^{\mathrm{traj}}\cup\mathcal{C}_e^{\mathrm{con}}$
    \State Group active cards in $\mathcal{C}_e$ into Evidence Windows $\{\mathcal{B}_k\}_{k=1}^{K_e}$
    \State $\{(u,\mathcal{C}(u)):u\in\mathcal{U}_e\}\gets\bigcup_{k=1}^{K_e}f^{\mathrm{edit}}(S_e^W,\mathcal{B}_k)$
    \For{$u\in\mathcal{U}_e$}
        \State $\mathcal{G}(u)\gets\bigcup_{C_i\in\mathcal{C}(u)}\mathcal{G}_i$
        \State Obtain source--replay pairs $\mathcal{T}_u$ by replaying $\mathcal{G}(u)$ under $S_e^W\oplus u$
        \State $(r_u,h_u)\gets f^{\mathrm{eval}}(S_e^W,u,\mathcal{C}(u),\mathcal{T}_u)$
        \If{$r_u=\mathrm{reflect}$}
            \State $u'\gets f^{\mathrm{edit}}(S_e^W,u,\mathcal{C}(u),\mathcal{T}_u,h_u)$
            \State $\mathcal{C}(u')\gets\mathcal{C}(u)$, $\mathcal{G}(u')\gets\mathcal{G}(u)$
            \State Obtain $\mathcal{T}_{u'}$ by replaying $\mathcal{G}(u')$ under $S_e^W\oplus u'$
            \State $(r_{u'},h_{u'})\gets f^{\mathrm{eval}}(S_e^W,u',\mathcal{C}(u'),\mathcal{T}_{u'})$
        \EndIf
    \EndFor
    \State $\mathcal{U}_e^{+}\gets\{u\in\mathcal{U}_e\mid r_u=\mathrm{accept}\}\cup\{u'\mid u\in\mathcal{U}_e,\ r_u=\mathrm{reflect},\ r_{u'}=\mathrm{accept}\}$
    \State $\overline{\mathcal{U}}_e\gets f^{\mathrm{edit}}(S_e^W,\mathcal{U}_e^{+})$, preserving supporting card associations
    \State $\widetilde{S}_e\gets S_e^W\oplus\overline{\mathcal{U}}_e$
    \If{$\widehat{J}_{\mathcal{D}_{\mathrm{val}}}(\widetilde{S}_e)>\widehat{J}_{\mathcal{D}_{\mathrm{val}}}(S_e^V)$}
        \State $S_{e+1}^V\gets\widetilde{S}_e$, $P_{e+1}\gets\varnothing$
        \State $\mathcal{Q}_{e+1}\gets P_e\cup\overline{\mathcal{U}}_e$
    \Else
        \State $S_{e+1}^V\gets S_e^V$
        \For{$v\in\overline{\mathcal{U}}_e$}
            \State Replay $\mathcal{G}(v)$ under $S_e^V\oplus v$ and assess with $f^{\mathrm{eval}}$ using $\mathcal{C}(v)$ to obtain $r_v^{\mathrm{post}}$
            \If{$r_v^{\mathrm{post}}=\mathrm{reflect}$}
                \State Revise $v$ into $v'$ with $f^{\mathrm{edit}}$ using $S_e^V$, supporting cards, replay pairs, and feedback
                \State $\mathcal{C}(v')\gets\mathcal{C}(v)$, $\mathcal{G}(v')\gets\mathcal{G}(v)$
                \State Replay $\mathcal{G}(v')$ under $S_e^V\oplus v'$ and assess with $f^{\mathrm{eval}}$ to obtain $r_{v'}^{\mathrm{post}}$
            \EndIf
        \EndFor
        \State $\mathcal{U}_e^{\mathrm{post},+}\gets\{v\in\overline{\mathcal{U}}_e\mid r_v^{\mathrm{post}}=\mathrm{accept}\}\cup\{v'\mid v\in\overline{\mathcal{U}}_e,\ r_v^{\mathrm{post}}=\mathrm{reflect},\ r_{v'}^{\mathrm{post}}=\mathrm{accept}\}$
        \State Form $P_{e+1}$ from $P_e$ by replacing targeted entries with edits in $\mathcal{U}_e^{\mathrm{post},+}$ and appending the remaining retained edits in order
        \State $\mathcal{Q}_{e+1}\gets P_{e+1}$
    \EndIf
    \State Update card lifecycle states from validation and replay outcomes to obtain $\mathcal{C}_{e+1}^{\mathrm{carry}}$
\EndFor
\State $S^\star\gets S_{E+1}^V$
\State \Return $S^\star$
\end{algorithmic}
\end{algorithm*}

At each epoch, \method{} forms the Working Skill $S_e^W=S_e^V\oplus P_e$ and collects training trajectories $\mathcal{R}_e$. The extractor produces Trajectory Evidence Cards $\mathcal{C}_e^{\mathrm{traj}}$ and Contrastive Evidence Cards $\mathcal{C}_e^{\mathrm{con}}$, using $\mathcal{Q}_e$ and its supporting cards to select adjacent-epoch trajectory pairs. Both sources join $\mathcal{C}_e^{\mathrm{carry}}$ in the Evidence Pool $\mathcal{C}_e$. Active cards are grouped into Evidence Windows $\mathcal{B}_k$ for synthesizing candidate edits $\mathcal{U}_e$ and their supporting associations $\mathcal{C}(u)$. Replay under $S_e^W\oplus u$, with one reflection opportunity, yields $\mathcal{U}_e^{+}$. These edits are consolidated into $\overline{\mathcal{U}}_e$ to form $\widetilde{S}_e$.

If global validation accepts $\widetilde{S}_e$, it becomes $S_{e+1}^V$ and $P_{e+1}$ is cleared. Otherwise, $S_e^V$ remains unchanged, and post-rejection replay evaluates each consolidated edit $v$ under $S_e^V\oplus v$. Accepted edits from post-rejection replay update $P_{e+1}$ for subsequent training and refinement. Card lifecycle updates determine $\mathcal{C}_{e+1}^{\mathrm{carry}}$, while $\mathcal{Q}_{e+1}$ identifies edits for further cross-epoch comparison. After $E$ epochs, the algorithm returns $S^\star=S_{E+1}^V$, excluding unpromoted provisional edits.
\subsection{Evidence Cards and Edit Representation}
\label{app:evidence_edit_representation}

A Replayable Evidence Card is represented by $C_i=\langle z_i,\mathcal{G}_i,\widetilde{u}_i\rangle$, where $z_i$ is a persistent Evidence identifier, $\mathcal{G}_i$ is a set of supporting trigger ranges, and $\widetilde{u}_i$ is the proposed correction. Each Card also stores its mechanism-level pattern, evidence source, outcome category, and optional failure type.

Each trigger range $g=(x,e,s,t)\in\mathcal{G}_i$ denotes the inclusive interval $[s,t]$ in the trajectory of training task $x$ collected at epoch $e$, so that the source state can be reconstructed. The range stores the task identifier, source epoch, step boundaries, actions, and environment observations. A Card may contain ranges from multiple tasks when they support the same correction.

We use two types of Cards: \textbf{Trajectory Evidence Cards}, extracted from current-epoch trajectories, and \textbf{Contrastive Evidence Cards}, generated by comparing trajectories associated with Tracked Edits across epochs. Contrastive Cards retain the identifier of the Origin Edit they assess. Before entering the Evidence Pool, each Card is checked by a grounding verifier against its referenced source trajectories, and Cards with invalid ranges or unsupported corrections are discarded.

Active Cards whose corrections are not already covered by the current skill are grouped into Evidence Windows for subsequent edit synthesis. For each Card, we construct a semantic representation from its pattern and proposed edit fields, including the operation, target, and content. Normalized embeddings of these representations are clustered with HDBSCAN~\citep{hdbscan} in the default implementation. Clustering is performed over edit semantics rather than complete trajectories or trigger ranges; trigger ranges are retained for replay. Oversized clusters are split according to the window capacity, and compatible small clusters are merged when possible.

A Card-level correction is represented as: 
\begin{equation}
\widetilde{u}_i=\langle a_i,t_i,c_i\rangle,
\end{equation}
where $a_i\in\{\texttt{append},\texttt{insert\_after},\texttt{insert\_after\_section},\texttt{replace},\texttt{delete}\}$ is the edit operation, $t_i$ is the target location or exact target text, and $c_i$ is the Markdown content to insert or replace. The target is empty only for \texttt{append}, and the content may be empty for \texttt{delete}.

Let $\mathcal{B}_k=\{C_i\mid i\in\mathcal{I}_k\}$ denote the $k$-th Evidence Window. The Evidence-Window editor refines or consolidates its Card-level corrections into an Evidence-Linked Edit:
\begin{equation}
u=\langle y_u,a_u,t_u,c_u,I_u,L_u\rangle,
\end{equation}
where $y_u$ is the edit identifier, $I_u$ is the set of supporting Card identifiers, and $L_u$ contains source edit identifiers when the edit is derived from previous edits. Its replay ranges are recovered as
\begin{equation}
\mathcal{G}(u)=\bigcup_{i:z_i\in I_u}\mathcal{G}_i.
\end{equation}
The identifiers in $I_u$ and $L_u$ are retained as provenance metadata and are not inserted into the external skill document. Card and edit lifecycle transitions are described in Section~\ref{app:evidence_edit_lifecycle}.
\subsection{Trigger-Range Replay}
\label{app:trigger_range_replay}

Trigger-Range Replay evaluates an Evidence-Linked Edit on the local behaviors cited by its supporting Evidence Cards. For each range $g=(x,e,s,t)\in\mathcal{G}(u)$, the implementation loads the recorded trajectory of task $x$ from epoch $e$ and replays the prefix before step $s$ to reconstruct the original environment state and interaction history. The Action Agent then executes the target interval using the candidate skill $S\oplus u$, allowing its actions to differ from the recorded trajectory. Prefix actions are used only for state reconstruction and are not counted as new agent decisions.

The replay record pairs the original and re-executed segments, including their actions, environment observations, progress signals, and terminal outcomes. Given these paired segments, the current skill, the candidate edit, and its supporting Cards, the replay evaluator returns $r_u\in\{\texttt{accept},\texttt{reflect},\texttt{reject}\}$ with feedback. An accepted edit is retained for subsequent consolidation into the current candidate revision; a reflected edit is revised while preserving its evidence links and evaluated again; and a rejected edit is excluded from the current candidate.

The same procedure is used at two stages. \textbf{Evidence-window replay} verifies constructed edits before consolidation, whereas \textbf{post-rejection replay} reassesses edits from a globally rejected revision before admitting them as Provisional Edits. Replay provides local behavioral verification, while global validation determines whether the consolidated candidate updates the Validated Skill.
\subsection{Evidence and Edit Lifecycle}
\label{app:evidence_edit_lifecycle}

\textbf{Evidence Card lifecycle.} A Card is generated from a current-epoch trajectory or a cross-epoch contrastive comparison and enters the Evidence Pool only after a grounding verifier confirms that its trigger ranges support the proposed pattern and correction. Verified Cards are initialized as \texttt{active} and may be selected into Evidence Windows across multiple epochs if their evidence remains unresolved. Cards supporting live Provisional Edits are marked as protected and reserved for cross-epoch comparison using adjacent-epoch trajectories, whereas Cards whose corrections are incorporated into a validated skill are marked as \texttt{archived}. Cards that are repeatedly rejected, ignored, or left without a usable edit may become \texttt{stale} after exceeding the configured rejection limit; unresolved active or deferred Cards remain available for future Evidence-Window construction.

\textbf{Edit lifecycle.} Card-level corrections are grouped into Evidence Windows and refined into Evidence-Linked Edits, which are checked for patch validity and evaluated through trigger-range replay. A replay returns \texttt{accept}, \texttt{reflect}, or \texttt{reject}: accepted edits are retained directly, reflected edits are rewritten as provenance-preserving successors, and rejected edits are excluded from the current revision unless recovered by later replay or cross-epoch correction. Accepted edits are consolidated into a candidate skill and evaluated on the development split. If validation succeeds, the changed edits become part of the Validated Skill and their historical records are preserved; if validation fails, the Validated Skill remains unchanged, while edits that pass post-rejection replay may enter the Provisional Edit Ledger with their evidence links intact and contribute to the next Working Skill. For a linked origin edit, a later Contrastive Evidence Card can generate a correction; corrections to Provisional Edits replace the live ledger entry, whereas corrections to validated edits are stored as new lineage-linked edits without modifying the original history.
\subsection{Implementation and Evaluation Details}
\label{app:implementation_details}

\paragraph{Evolution schedule and rollout.} \method{} performs four skill-evolution epochs. Each epoch traverses the complete training task set $\mathcal{D}_{\mathrm{tr}}$ once, with the task order shuffled at the beginning of the epoch and no sampling with replacement. The Action Agent $\pi$ executes tasks under the Working Skill $S_e^W$ in rollout bundles of four using four isolated environment workers, with one trajectory per task. The LLM evidence extractor $f^{\mathrm{ext}}$ processes the resulting trajectories $\mathcal{R}_e$ in minibatches of four and produces at most four card-level correction proposals per minibatch.

\paragraph{Evidence grouping and skill evolution.} We encode each active Replayable Evidence Card $C_i$ with the local \texttt{Qwen3-Embedding-0.6B} encoder~\citep{qwen3embedding} using a batch size of 32 and group the resulting embeddings with HDBSCAN under Euclidean distance. We set \texttt{min\_cluster\_size}=4 and \texttt{min\_samples}=2. If the number of active cards in the Evidence Pool $\mathcal{C}_e$ is smaller than either setting, the corresponding value is clipped to that number. HDBSCAN~\citep{hdbscan} noise points are retained as singleton groups rather than discarded. After clustering, Evidence Windows $\mathcal{B}_k$ contain 4--20 cards. When necessary, small components are merged or rebalanced using a centroid cosine-similarity threshold of $0.58$. The implementation uses \texttt{cluster\_step}, and all resulting windows are processed without an additional fixed limit on $K_e$. For each $\mathcal{B}_k$, the LLM editor $f^{\mathrm{edit}}$ uses a candidate-budget hint of five edits. Replay-verified edits in $\mathcal{U}_e^{+}$ are consolidated in chunks of at most eight candidates before a global consolidation pass, with an epoch-level budget of 128 edits. Each candidate edit $u$ is replayed on at most three trigger ranges from $\mathcal{G}(u)$. Post-rejection replay evaluates at most 15 consolidated edit instances under $S_e^V\oplus v$ in each non-final epoch and is disabled after the final validation gate. The Provisional Edit Ledger $P_e$ contains at most 16 live edits. A card $C_i$ becomes stale after two explicit replay rejections and may remain deferred for at most two distinct epochs. For cross-epoch comparison, at most 20 task-level transition lineages associated with the Tracked Edits $\mathcal{Q}_e$ are retained for the next epoch.

\paragraph{Models and environments.} The Action Agent $\pi$ uses GPT-5.5, GPT-5.4, GPT-5.4-mini, Qwen3.5-4B, Qwen3.5-9B, or Qwen3.6-35B-A3B. GPT backbones use medium reasoning, while Qwen backbones use their non-reasoning configuration. The evidence extractor $f^{\mathrm{ext}}$, editor $f^{\mathrm{edit}}$, and evaluator $f^{\mathrm{eval}}$ use GPT-5.5 with medium reasoning. The temperature of $\pi$ is $0.7$ across all six backbones. ALFWorld and AppWorld permit at most 50 environment steps per task. ScienceWorld uses its task-specific recommended limit when available and a fallback limit of 100 steps otherwise. In all three environments, $\pi$ receives the eight most recent interaction steps.

\section{Preliminary Study Details}
\label{app:preliminary_details}
\vspace{-5pt}
\subsection{Trajectory-Derived Edit Outcomes}
\label{app:trajectory_edit_outcomes}

We examine edits generated by Trace2Skill from training trajectories on ALFWorld, AppWorld, and ScienceWorld. Each edit $u$ is associated with a set of training tasks $T(u)$ that motivated it, and the size of $T(u)$ varies across edits. We assess each edit independently by applying it to the skill version immediately preceding its generation, without applying other generated edits. The Action Agent then reruns every task in $T(u)$ from its initial state under the edited skill until the task terminates or reaches its step limit. We compare task-success accuracy on $T(u)$ with the accuracy of the corresponding pre-edit skill on the same tasks. An edit is classified as \emph{accept} if accuracy strictly increases, \emph{reject} if it strictly decreases, and \emph{reflect} if it remains unchanged.

\vspace{-5pt}
\subsection{Useful Edits within Rejected Revisions}
\label{app:rejected_revision_analysis}

We analyze rejected revisions produced by Trace2Skill. For each revision, Trace2Skill generates $N$ edits from training trajectories, with each edit associated with its source training task. The complete revision is rejected when its validation accuracy does not strictly exceed that of the previous Validated Skill. To examine whether the revision contains a useful subset of edits, we provide GPT-5.5 with the $N$ edit records and ask it to return the indices of the $\min(3,N-2)$ edits judged most promising based on their content. The selector does not rewrite the edits. We apply all selected edits together to the previous Validated Skill and evaluate the resulting skill with GPT-5.4-mini on the full training set. We then compare its task-success accuracy with that of the previous Validated Skill on the same tasks. The analysis shows that the selected edit sets improve training accuracy despite the rejection of their complete revisions on the validation set. This result shows that revision-level rejection can discard a combination of edits with measurable benefits.
\vspace{-5pt}

\section{Additional Experimental Results}
\label{app:additional_results}
\subsection{Complete Main Results}
\label{app:complete-main-results}

\providecommand{\method}{\textsc{EviSkill}}
\providecommand{\bestresult}[1]{\textbf{#1}}
\providecommand{\secondresult}[1]{\underline{#1}}
\definecolor{eviskillblue}{RGB}{232,242,255}

\begin{table}[t]
\caption{Task-success accuracy (\%) across three interactive benchmarks. $\Delta_{\mathrm{NS}}$ denotes the percentage-point change relative to No skill for the same model and dataset. \textbf{Bold} and \underline{underlined} values indicate the best and second-highest distinct accuracies within each model--dataset setting.}
\label{tab:complete-main-results}
\centering
\fontsize{6.0}{6.8}\selectfont
\setlength{\tabcolsep}{1.0pt}
\renewcommand{\arraystretch}{1.02}
\begin{tabularx}{\columnwidth}{@{}
>{\hsize=1.35\hsize\raggedright\arraybackslash}X
>{\hsize=.95\hsize\raggedright\arraybackslash}X
*{6}{>{\hsize=.95\hsize\raggedleft\arraybackslash}X}@{}}
\toprule
\multirow[c]{2}{=}{Model} &
\multirow[c]{2}{=}{Method} &
\multicolumn{2}{c}{ALFWorld (134)} &
\multicolumn{2}{c}{AppWorld (168)} &
\multicolumn{2}{c}{ScienceWorld (211)} \\
\cmidrule(lr){3-4}\cmidrule(lr){5-6}\cmidrule(lr){7-8}
& & Acc.$\uparrow$ & $\Delta_{\mathrm{NS}}$ &
Acc.$\uparrow$ & $\Delta_{\mathrm{NS}}$ &
Acc.$\uparrow$ & $\Delta_{\mathrm{NS}}$ \\
\midrule
\multirow[c]{7}{=}{GPT-5.5}
& No Skill & 90.30 & 0.00 & 85.71 & 0.00 & 76.30 & 0.00 \\
& LLM Skill & 94.03 & +3.73 & \secondresult{94.05} & +8.33 & 76.78 & +0.47 \\
& Trace2Skill & 94.03 & +3.73 & 86.31 & +0.60 & 65.40 & -10.90 \\
& EvoSkill & 94.78 & +4.48 & 91.07 & +5.36 & 54.50 & -21.80 \\
& SkillGrad & \secondresult{95.52} & +5.22 & 86.31 & +0.60 & 67.30 & -9.00 \\
& SkillOpt & 94.78 & +4.48 & 92.86 & +7.14 & \secondresult{80.09} & +3.79 \\
\rowcolor{eviskillblue}
& \method{} & \bestresult{96.27} & +5.97 & \bestresult{94.64} & +8.93 & \bestresult{83.41} & +7.11 \\
\midrule
\addlinespace[1.0pt]
\multirow[c]{7}{=}{GPT-5.4}
& No Skill & 75.37 & 0.00 & 88.10 & 0.00 & 52.13 & 0.00 \\
& LLM Skill & 89.55 & +14.18 & 88.69 & +0.60 & \secondresult{66.35} & +14.22 \\
& Trace2Skill & \secondresult{94.78} & +19.40 & 85.71 & -2.38 & 56.40 & +4.27 \\
& EvoSkill & 90.30 & +14.93 & 79.17 & -8.93 & 63.03 & +10.90 \\
& SkillGrad & 94.03 & +18.66 & 83.93 & -4.17 & 53.55 & +1.42 \\
& SkillOpt & \secondresult{94.78} & +19.40 & \secondresult{89.29} & +1.19 & 63.03 & +10.90 \\
\rowcolor{eviskillblue}
& \method{} & \bestresult{95.52} & +20.15 & \bestresult{90.48} & +2.38 & \bestresult{76.78} & +24.64 \\
\midrule
\addlinespace[1.0pt]
\multirow[c]{7}{=}{GPT-5.4-mini}
& No Skill & 74.63 & 0.00 & 82.14 & 0.00 & 38.86 & 0.00 \\
& LLM Skill & 88.06 & +13.43 & 76.19 & -5.95 & 36.97 & -1.90 \\
& Trace2Skill & 82.84 & +8.21 & 83.33 & +1.19 & 42.65 & +3.79 \\
& EvoSkill & 86.57 & +11.94 & \bestresult{85.12} & +2.98 & 44.55 & +5.69 \\
& SkillGrad & 88.06 & +13.43 & 83.33 & +1.19 & 44.08 & +5.21 \\
& SkillOpt & \secondresult{89.55} & +14.93 & \secondresult{84.52} & +2.38 & \secondresult{58.29} & +19.43 \\
\rowcolor{eviskillblue}
& \method{} & \bestresult{91.79} & +17.16 & 83.93 & +1.79 & \bestresult{67.78} & +28.91 \\
\midrule
\addlinespace[1.0pt]
\multirow[c]{7}{=}{Qwen3.5-4B}
& No Skill & 43.28 & 0.00 & 18.45 & 0.00 & 4.27 & 0.00 \\
& LLM Skill & 75.37 & +32.09 & 28.57 & +10.12 & 3.79 & -0.47 \\
& Trace2Skill & 82.84 & +39.55 & \bestresult{32.14} & +13.69 & 9.48 & +5.21 \\
& EvoSkill & 73.88 & +30.60 & 27.98 & +9.52 & 2.84 & -1.42 \\
& SkillGrad & \secondresult{84.33} & +41.04 & 30.36 & +11.90 & 6.64 & +2.37 \\
& SkillOpt & 83.58 & +40.30 & \secondresult{30.95} & +12.50 & \secondresult{14.69} & +10.43 \\
\rowcolor{eviskillblue}
& \method{} & \bestresult{85.07} & +41.79 & \bestresult{32.14} & +13.69 & \bestresult{20.85} & +16.59 \\
\midrule
\addlinespace[1.0pt]
\multirow[c]{7}{=}{Qwen3.5-9B}
& No Skill & 54.48 & 0.00 & 25.60 & 0.00 & 8.53 & 0.00 \\
& LLM Skill & 76.12 & +21.64 & 43.45 & +17.86 & 9.00 & +0.47 \\
& Trace2Skill & 75.37 & +20.90 & 45.24 & +19.64 & 19.91 & +11.37 \\
& EvoSkill & 76.87 & +22.39 & 32.74 & +7.14 & 17.06 & +8.53 \\
& SkillGrad & \secondresult{87.31} & +32.84 & \secondresult{46.43} & +20.83 & 14.22 & +5.69 \\
& SkillOpt & 86.57 & +32.09 & \bestresult{47.62} & +22.02 & \secondresult{22.27} & +13.74 \\
\rowcolor{eviskillblue}
& \method{} & \bestresult{91.04} & +36.57 & 43.45 & +17.86 & \bestresult{37.91} & +29.38 \\
\midrule
\addlinespace[1.0pt]
\multirow[c]{7}{=}{\shortstack[l]{Qwen3.6-35B-\\A3B}}
& No Skill & 70.90 & 0.00 & 54.17 & 0.00 & 17.06 & 0.00 \\
& LLM Skill & 79.85 & +8.96 & 58.33 & +4.17 & 17.54 & +0.47 \\
& Trace2Skill & 85.07 & +14.18 & \bestresult{64.88} & +10.71 & 25.12 & +8.06 \\
& EvoSkill & 79.85 & +8.96 & 55.36 & +1.19 & 23.22 & +6.16 \\
& SkillGrad & 85.82 & +14.93 & 60.71 & +6.55 & 19.91 & +2.84 \\
& SkillOpt & \secondresult{86.57} & +15.67 & 62.50 & +8.33 & \bestresult{42.18} & +25.12 \\
\rowcolor{eviskillblue}
& \method{} & \bestresult{88.06} & +17.16 & \secondresult{63.10} & +8.93 & \secondresult{40.76} & +23.70 \\
\bottomrule
\end{tabularx}
\vspace{1.5pt}
\end{table}

Table~\ref{tab:complete-main-results} reports the complete Q1 results across all six Action-Agent backbones, comprising three GPT backbones and three Qwen backbones. In addition to the four representative backbones reported in Table~\ref{tab:main-results}, the complete comparison includes GPT-5.4-mini and Qwen3.6-35B-A3B.

The benchmark-wise results reveal different levels of consistency across environments. On ALFWorld, \method{} achieves the highest accuracy with all six backbones. On ScienceWorld, it ranks first with five backbones and second with Qwen3.6-35B-A3B. Performance is more variable on AppWorld: \method{} ranks first or tied first in three settings and second in one, while continuing to outperform No skill in the remaining two. This distribution indicates that the relative advantage of \method{} is highly consistent on ALFWorld and ScienceWorld, whereas AppWorld presents greater variation among skill-evolution methods across the evaluated backbones.

The magnitude of improvement also varies across model families. Averaged over the nine model--dataset settings within each family, \method{} improves over No skill by 13.00 percentage points for the GPT backbones and 22.85 points for the Qwen backbones. The larger absolute gain for the Qwen family should be interpreted alongside its generally lower No-skill accuracies, which leave greater room for improvement. Overall, the complete results show that the gains of \method{} extend across backbone families and model scales, rather than depending on a particular Action Agent.

\subsection{Complete Ablation Results}
\label{app:complete_ablation_results}

\begin{figure*}[t]
\centering
\begin{subfigure}[t]{0.48\textwidth}
\centering
\includegraphics[width=\linewidth]{Figure/gpt_family_ablation.pdf}
\caption{Average results across the three GPT backbones.}
\label{fig:complete_ablation_gpt}
\end{subfigure}
\hfill
\begin{subfigure}[t]{0.48\textwidth}
\centering
\includegraphics[width=\linewidth]{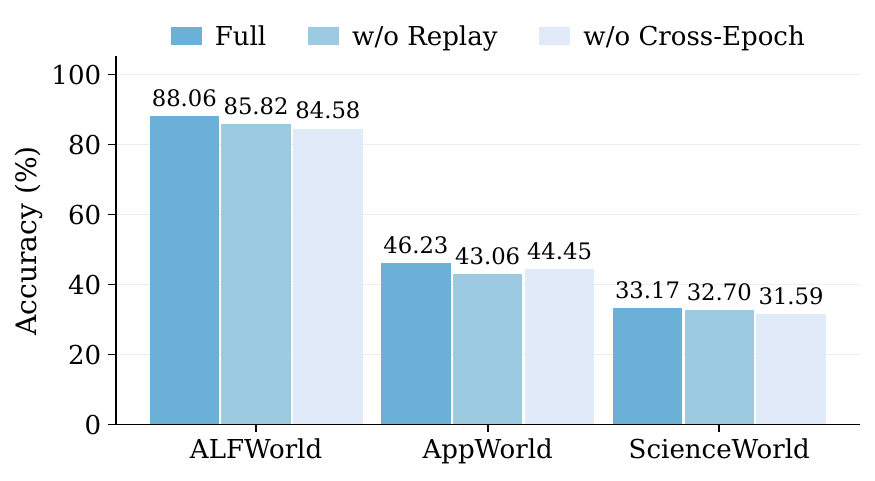}
\caption{Average results across the three Qwen backbones.}
\label{fig:complete_ablation_qwen}
\end{subfigure}
\caption{Family-level component ablations across all six Action-Agent backbones. Each panel reports task-success accuracy averaged over the three backbones in the corresponding model family. w/o Replay removes Replay-Guided Edit Verification, while w/o Cross-Epoch disables Cross-Epoch Evidence Propagation.}
\label{fig:complete_ablation_results}
\end{figure*}

Figure~\ref{fig:complete_ablation_results} extends the ablation study in Section~\ref{sec:ablation} to all six Action-Agent backbones. For each benchmark, the GPT results are averaged over GPT-5.5, GPT-5.4, and GPT-5.4-mini, while the Qwen results are averaged over Qwen3.5-4B, Qwen3.5-9B, and Qwen3.6-35B-A3B. Full denotes the complete \method{} framework; w/o Replay removes Replay-Guided Edit Verification, and w/o Cross-Epoch disables Cross-Epoch Evidence Propagation. We first examine the family-level effects of these two components and then further separate the two persistent pathways within Cross-Epoch Evidence Propagation.

\paragraph{Effects of key components.} Full \method{} achieves the highest accuracy across all benchmarks in both model families. For GPT backbones (Figure~\ref{fig:complete_ablation_gpt}), removing replay causes larger drops than removing cross-epoch propagation on all three benchmarks, with the largest decreases on AppWorld (4.56 and 4.16 percentage points, respectively). For Qwen backbones (Figure~\ref{fig:complete_ablation_qwen}), cross-epoch propagation contributes more on ALFWorld and ScienceWorld, with drops of 3.48 and 1.58 points, compared with 2.24 and 0.47 points without replay. On AppWorld, removing replay causes the larger drop (3.17 versus 1.78 points). These results support both components while showing that their relative contributions depend on the backbone family and environment.

\paragraph{Decomposition of Cross-Epoch Evidence Propagation.}

\begin{wraptable}{r}{0.44\textwidth}
\centering
\small
\vspace{5pt}
\caption{Cross-epoch ablations.}
\vspace{-10pt}
\label{tab:cross_epoch_decomposition}
\renewcommand{\arraystretch}{1.08}
\setlength{\tabcolsep}{4pt}
\begin{tabular*}{\linewidth}{@{\extracolsep{\fill}}lcc@{}}
\toprule
Variant & Accuracy (\%) & Gain \\
\midrule
Full & \textbf{88.06} & +3.48 \\
Cards Only & 86.32 & +1.74 \\
Edits Only & 86.82 & +2.24 \\
w/o Cross-Epoch & 84.58 & 0.00 \\
\bottomrule
\end{tabular*}
\end{wraptable}
We separate two cross-epoch pathways: carrying unresolved cards into subsequent Evidence Pools and retaining replay-supported provisional edits in subsequent Working Skills. We evaluate on ALFWorld, where removing cross-epoch propagation causes the largest Qwen-family decrease, and average accuracies over Qwen3.5-4B, Qwen3.5-9B, and Qwen3.6-35B-A3B. Cards Only preserves unresolved cards but discards provisional edits after rejected revisions. Edits Only retains provisional edits but excludes unresolved cards from the next Evidence Pool. Table~\ref{tab:cross_epoch_decomposition} reports accuracy and percentage-point gains over w/o Cross-Epoch. Cards Only and Edits Only improve accuracy by 1.74 and 2.24 points, respectively. Full achieves 88.06\% accuracy, exceeding the two variants by 1.74 and 1.24 points. Both pathways contribute to performance, and combining them provides further gains over retaining either alone.

\subsection{Contrastive Evidence-Guided Correction of Tracked Edits}
\label{app:contrastive_rewrite}

We examine whether cross-epoch contrastive evidence leads to replay-supported corrections of the Tracked Edits it targets. Following Section~\ref{sec:edit_construction}, the supporting cards of $\mathcal{Q}_e$ identify tasks for adjacent-epoch trajectory comparison. The extractor uses these comparisons to identify persistent deficiencies or newly exposed problems and produces Contrastive Evidence Cards linked to the relevant edits. These cards guide targeted edit synthesis, followed by replay verification and, when needed, reflection.

\begin{wrapfigure}{r}{0.46\textwidth}
    \centering
    \includegraphics[width=\linewidth]{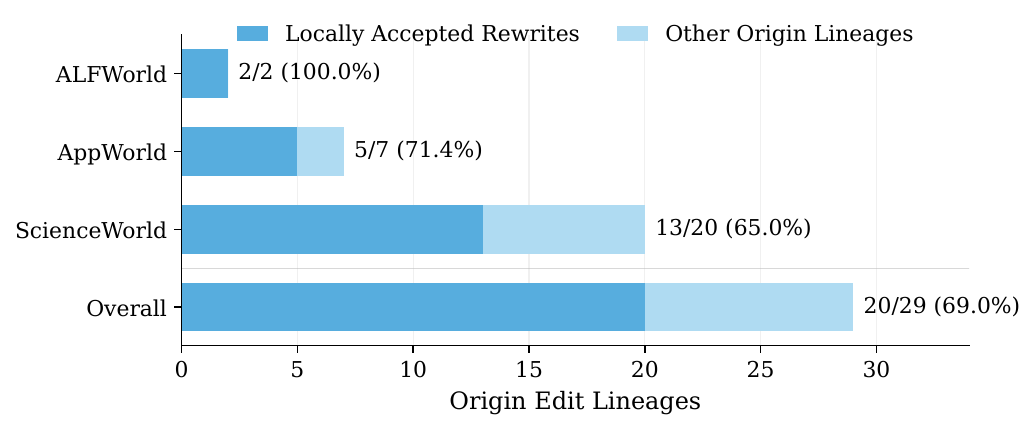}
    \caption{Replay-supported correction of origin edit lineages.}
    \label{fig:contrastive_rewrite_conversion}
    \vspace{-10pt}
\end{wrapfigure}

Across 18 complete runs, 48 Contrastive Evidence Cards target 29 distinct original edits. For Figure~\ref{fig:contrastive_rewrite_conversion}, we group linked corrections by their original edit, forming an \emph{origin edit lineage}. Each lineage is counted once: dark blue denotes at least one replay-accepted correction, and light blue denotes none. The bar length gives the number of targeted lineages, while each annotation reports the accepted count, total count, and their ratio. 

Overall pools the three benchmarks. Of the 29 lineages, 20 (69.0\%) yield a replay-accepted correction. ALFWorld achieves 2/2, although this covers only two lineages. AppWorld achieves 5/7 (71.4\%), and ScienceWorld achieves 13/20 (65.0\%). ScienceWorld accounts for most targeted lineages and accepted corrections, but also seven of the nine lineages without an accepted correction.
These results connect evidence propagation to behavioral verification: cross-epoch comparisons supply actionable evidence for revising existing guidance, and replay supports corrections for a majority of the targeted lineages. The nine remaining lineages also show that identifying a deficiency does not ensure an effective repair. Acceptance here establishes local replay support, while incorporation into the Validated Skill remains subject to global validation.
\subsection{Evidence Support Breadth and Granularity}
\label{app:evidence_support_breadth}

We examine whether evidence supported by multiple tasks is more likely to support edits that survive replay and contribute to subsequent skill revisions. This analysis tests two aspects of our design: whether broader evidence support can substitute for behavioral verification, and whether cross-task evidence remains useful after verification. A Replayable Evidence Card is classified as \emph{cross-task} if its trigger ranges reference at least two distinct training tasks, and \emph{single-task} otherwise. Across 18 complete runs, the 3,313 Trajectory Evidence Cards comprise 2,450 single-task cards and 863 cross-task cards (26.05\%). We restrict this analysis to trajectory-derived cards to examine task breadth within one extraction source, excluding the 48 Contrastive Evidence Cards.

Table~\ref{tab:evidence_breadth_utility} follows card--edit associations through synthesis, replay, and downstream skill updates. Edit-linked measures participation in candidate edit synthesis. Replay-retained reports retention among cards associated with replayed edits. Final Provisional and Validated report associations with edits remaining provisional at the end of evolution or incorporated into the Validated Skill, respectively. The final column measures the proportion of replay-retained cards associated with either downstream state. These are card-level provenance statistics, and a card may support multiple edits.

\begin{table*}[!htbp]
  \caption{Card utilization by support breadth. Percentages use the denominators specified in the column headers.}
  \label{tab:evidence_breadth_utility}
  \centering
  \scriptsize
  \setlength{\tabcolsep}{4.1pt}
  \renewcommand{\arraystretch}{1.08}
  \begin{tabular*}{\textwidth}{@{\extracolsep{\fill}}llrccccc@{}}
    \toprule
    Dataset & Support & Cards &
    \shortstack{Edit-linked\\(of Cards)} &
    \shortstack{Replay-retained\\(of replayed)} &
    \shortstack{Final Provisional\\(of Cards)} &
    \shortstack{Validated\\(of Cards)} &
    \shortstack{Prov./validated\\(of retained)} \\
    \midrule
    ALFWorld & Single-task & 532 & 151 (28.38) & 139/151 (92.05) & 30 (5.64) & 92 (17.29) & 122/139 (87.77) \\
             & Cross-task  & 389 & 154 (39.59) & 136/154 (88.31) & 17 (4.37) & 110 (28.28) & 127/136 (93.38) \\
    \addlinespace[2pt]
    AppWorld & Single-task & 695 & 580 (83.45) & 451/570 (79.12) & 32 (4.60) & 347 (49.93) & 379/451 (84.04) \\
             & Cross-task  & 123 & 87 (70.73) & 59/85 (69.41) & 3 (2.44) & 53 (43.09) & 55/59 (93.22) \\
    \addlinespace[2pt]
    ScienceWorld & Single-task & 1,223 & 1,182 (96.65) & 818/1,159 (70.58) & 0 (0.00) & 732 (59.85) & 730/818 (89.24) \\
                 & Cross-task  & 351 & 343 (97.72) & 234/338 (69.23) & 0 (0.00) & 212 (60.40) & 211/234 (90.17) \\
    \bottomrule
  \end{tabular*}
\end{table*}

\textbf{Broader evidence supports that replay does not guarantee retention.}
Cross-task cards have lower replay-retention rates than single-task cards on all three benchmarks: 88.31\% versus 92.05\% on ALFWorld, 69.41\% versus 79.12\% on AppWorld, and 69.23\% versus 70.58\% on ScienceWorld. Observing a shared pattern across tasks therefore does not ensure that the synthesized edit produces its intended effect. This supports the separation between Evidence-Grounded Edit Synthesis and Replay-Guided Edit Verification.

\textbf{Replay-retained cross-task evidence has higher downstream utilization.}
Among replay-retained cards, cross-task cards more frequently support provisional or validated edits: 93.38\% versus 87.77\% on ALFWorld, 93.22\% versus 84.04\% on AppWorld, and 90.17\% versus 89.24\% on ScienceWorld. This consistent pattern suggests that broader support can remain useful after behavioral verification, motivating the preservation of card--edit associations through subsequent skill updates.

\textbf{The value of support breadth depends on the environment.}
Cross-task cards have higher edit-linking and validated-association rates on ALFWorld, lower rates on AppWorld, and similar rates on ScienceWorld. Single-task evidence therefore remains a substantial source of useful guidance. These results support retaining both forms of evidence and assessing their resulting edits through replay, without treating task breadth alone as a quality criterion or evidence of generalization to unseen tasks.

\paragraph{Supporting trigger ranges.}
We measure the amount of execution evidence associated with each Replayable Evidence Card $C_i$ by summing the lengths of its supporting trigger ranges $\mathcal{G}_i$:
\begin{equation}
L(C_i)=\sum_{g=(x,e_g,s,t)\in\mathcal{G}_i}(t-s+1).
\label{eq:card_trigger_range_length}
\end{equation}
The endpoints are inclusive, and overlapping ranges are counted separately. Unlike the preceding task-breadth analysis, this analysis includes both Trajectory Evidence Cards and Contrastive Evidence Cards, which share the same trigger-range representation.

\begin{wraptable}{r}{0.46\textwidth}
\centering
\small
\caption{Supporting trigger-range statistics.}
\label{tab:supporting_trigger_ranges}
\setlength{\tabcolsep}{3pt}
\renewcommand{\arraystretch}{1.05}
\begin{tabular*}{\linewidth}{@{\extracolsep{\fill}}lrrrr@{}}
\toprule
Dataset & Cards & Ranges & Steps & Steps/Card \\
\midrule
ALFWorld     & 923   & 1,482 & 8,160  & 8.84  \\
AppWorld     & 828   & 1,035 & 5,520  & 6.67  \\
ScienceWorld & 1,610 & 2,142 & 20,093 & 12.48 \\
\midrule
Overall      & 3,361 & 4,659 & 33,773 & 10.05 \\
\bottomrule
\vspace{-15pt}
\end{tabular*}
\end{wraptable}
Table~\ref{tab:supporting_trigger_ranges} reports card and trigger-range counts, summed range lengths (Steps), and average lengths per card (Steps/Card). Overall, the three benchmarks pool to 3,361 cards and 4,659 ranges spanning 33,773 counted steps, averaging 10.05 steps per card. The median is 7 steps, and 1,057 cards (31.45\%) contain multiple ranges.
ScienceWorld has the longest supporting evidence per card at 12.48 steps, followed by ALFWorld at 8.84 and AppWorld at 6.67. By extraction source, the 3,313 Trajectory Evidence Cards average 10.08 steps per card, compared with 7.63 for the 48 Contrastive Evidence Cards.
\subsection{Cross-Epoch Edit Retention and Evidence Reuse}
\label{app:cross_epoch_reuse}

This section examines how replay-supported edits from globally rejected revisions are retained and subsequently incorporated into the Validated Skill, and how unresolved cards are reused across epochs.

\begin{figure*}[t]
    \centering
    \begin{subfigure}[t]{0.46\textwidth}
        \centering
        \includegraphics[width=\linewidth]{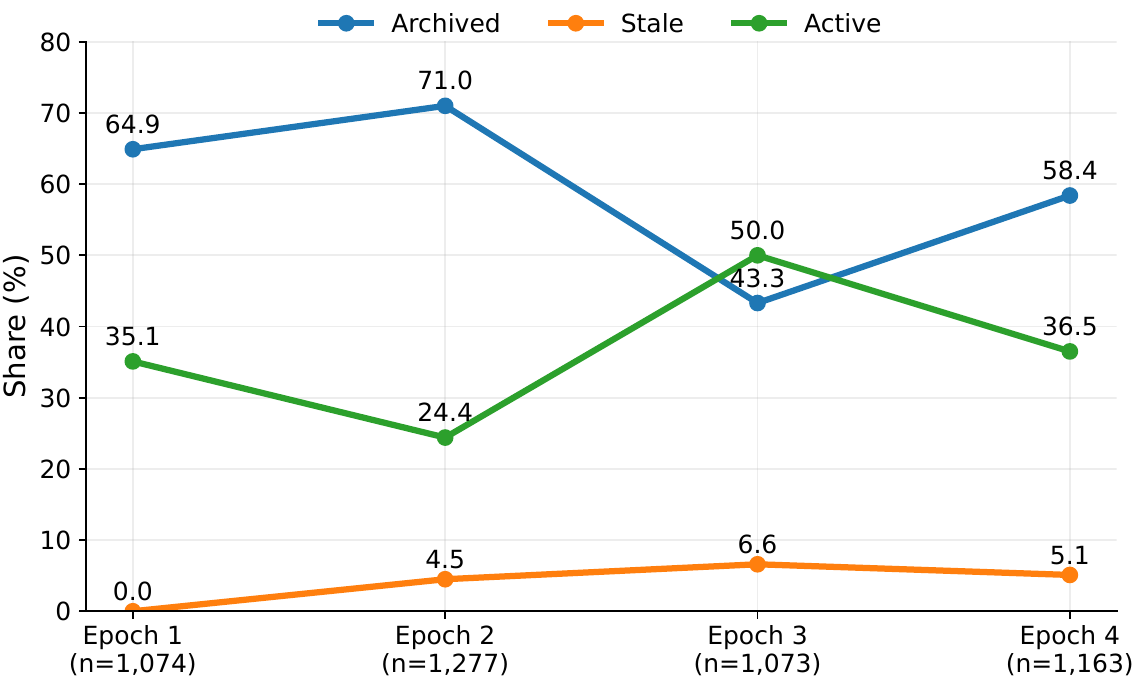}
        \caption{Evidence Pool lifecycle composition.}
        \label{fig:app_evidence_pool}
    \end{subfigure}
    \hfill
    \begin{subfigure}[t]{0.52\textwidth}
        \centering
        \includegraphics[width=\linewidth]{Figure/evidence_lifecycle_b.pdf}
        \caption{Evidence-Window assignment depth across epochs.}
        \label{fig:app_evidence_depth}
    \end{subfigure}
    \caption{Cross-epoch lifecycle of Evidence Cards over four skill-evolution epochs. (a) Lifecycle-state shares among Cards processed from each epoch's input Evidence Pool, pooled across 18 runs; \(n\) gives the input pool size. (b) Distribution of unique Evidence Cards by the number of distinct epochs in which they are assigned to an Evidence Window. Pre-resolved denotes Cards resolved without any Evidence-Window assignment. Percentages are computed within each benchmark, and All is the count-weighted aggregate.}
    \label{fig:app_evidence_lifecycle}
\end{figure*}

\paragraph{Evidence Pool lifecycle.} Figure~\ref{fig:app_evidence_pool} shows how the Cards in each epoch's input Evidence Pool are distributed among the Archived, Stale, and Active states. Archived Cards support corrections that have been incorporated into an accepted skill revision and are excluded from subsequent Evidence-Window construction. Active Cards remain eligible for later Evidence Windows, whereas Active Cards whose associated edit proposals are repeatedly rejected during replay-guided verification are marked as Stale and excluded from further Window construction.

The input pool contains 1,074, 1,277, 1,073, and 1,163 Cards in Epochs~1--4, respectively. Archived Cards account for 64.9\%, 71.0\%, 43.3\%, and 58.4\% of these pools, while Active Cards account for 35.1\%, 24.4\%, 50.0\%, and 36.5\%. The Stale share remains limited, increasing from 0.0\% in Epoch~1 to 4.5\%, 6.6\%, and 5.1\% in the subsequent epochs. The non-monotonic pool sizes and changing lifecycle composition show that the Evidence Pool is continually updated through retention, archiving, and staleness rather than simply accumulated across epochs.

\paragraph{Cross-epoch Evidence-Window assignment.} Figure~\ref{fig:app_evidence_depth} measures the number of distinct epochs in which each Evidence Card is assigned to an Evidence Window. The analysis covers 3,361 unique Cards, including 923 from ALFWorld, 828 from AppWorld, and 1,610 from ScienceWorld. Overall, 22.7\% are pre-resolved without an Evidence-Window assignment, and 54.3\% are assigned in exactly one epoch. The remaining 23.0\% are assigned across multiple epochs: 20.3\% in two epochs, 2.0\% in three, and 0.7\% in all four.

The assignment patterns vary across benchmarks. In ALFWorld, 66.6\% of Cards are pre-resolved, 31.1\% are assigned in one epoch, and only 2.3\% are assigned in two. In AppWorld, 52.7\% are assigned in one epoch, while 28.0\%, 1.3\%, and 0.2\% are assigned in two, three, and four epochs, respectively. ScienceWorld contains no pre-resolved Cards in this accounting; 68.4\% are assigned in one epoch, and 26.6\%, 3.5\%, and 1.4\% are assigned in two, three, and four epochs. Thus, most Cards participate in Evidence-Window construction within a single epoch, while a substantial subset is reassigned across multiple epochs as unresolved evidence remains available.

\begin{wrapfigure}{r}{0.44\textwidth}
    \centering
    \includegraphics[width=\linewidth]{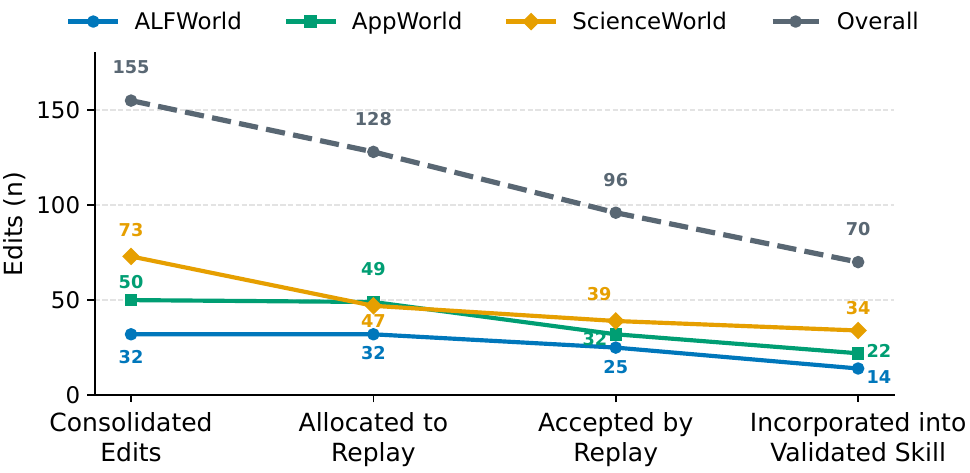}
    \caption{Post-rejection replay and subsequent edit incorporation.}
    \label{fig:app_cross_epoch_retention}
    \vspace{-10pt}
\end{wrapfigure}
\paragraph{Post-rejection replay and subsequent incorporation.}
Figure~\ref{fig:app_cross_epoch_retention} traces consolidated edits from globally rejected revisions through replay allocation, replay acceptance, and incorporation into the Validated Skill through later revisions that pass global validation. Overall pools counts across the three benchmarks and 18 runs. Of 155 consolidated edits, 128 undergo post-rejection replay, 96 are accepted, and 70 are subsequently incorporated. Thus, replay retains 75.0\% of the allocated edits, and 72.9\% of those accepted later enter the Validated Skill. Global rejection therefore does not eliminate the potential value of individual edits.
The largest reduction before replay occurs on ScienceWorld, where 47 of 73 edits are allocated, compared with 32 of 32 on ALFWorld and 49 of 50 on AppWorld. ScienceWorld nevertheless has the highest replay acceptance rate (39/47, 83.0\%) and subsequent incorporation rate among accepted edits (34/39, 87.2\%). The latter rate is 56.0\% on ALFWorld (14/25) and 68.8\% on AppWorld (22/32). These differences show that both replay retention and later incorporation vary across environments, with ScienceWorld contributing the most incorporated edits despite its smaller allocation fraction.

\section{Efficiency Analysis}
\label{app:efficiency}


\subsection{Validation Gain per Training Rollout}
\label{app:validation_gain_per_rollout}

We report validation accuracy improvement normalized by the number of training
rollouts during skill evolution. For each dataset--backbone cell $c$, we compute
\begin{equation}
  E_{100}^{(c)} =
  \frac{A_{\mathrm{val},c}^{\mathrm{final}}-
        A_{\mathrm{val},c}^{\mathrm{initial}}}
       {N_{\mathrm{train},c}}\times 100,
  \label{eq:validation_gain_per_100_rollouts}
\end{equation}
where $A_{\mathrm{val}}$ denotes validation accuracy used for global validation
and skill selection, and $N_{\mathrm{train},c}$ is the number of training-task
rollouts consumed during skill evolution.

\begin{table*}[!htbp]
  \caption{Validation gain per 100 training rollouts on the common 18-cell subset. ``Mean rollouts'' and ``mean gain'' are cell-macro averages; gain is measured in percentage points (pp). Dataset and overall columns report $E_{100}$ in validation-accuracy points per 100 training rollouts. Bold denotes the highest value in each comparison column.}
  \label{tab:validation_gain_per_rollout}
  \centering
  \small
  \setlength{\tabcolsep}{5pt}
  \renewcommand{\arraystretch}{1.08}
  \begin{tabular*}{\textwidth}{@{\extracolsep{\fill}}lrrrrrr@{}}
    \toprule
    Method & \shortstack{Mean train\\rollouts} &
    \shortstack{Mean val.\\gain (pp)} &
    ALFWorld & AppWorld & ScienceWorld & Overall \\
    \midrule
    \method{} & 376.00 & \textbf{14.25} & \textbf{1.99} & \textbf{3.01} & \textbf{6.33} & \textbf{3.77} \\
    EvoSkill  & 376.00 & 5.86 & 0.18 & 1.22 & 3.26 & 1.55 \\
    SkillGrad & 119.94 & 1.18 & 0.07 & 0.91 & 1.53 & 0.84 \\
    \bottomrule
  \end{tabular*}
\end{table*}

\method{} obtains 3.77 validation-accuracy points per 100 training rollouts on all metrics, compared with 1.55 for EvoSkill and 0.84 for SkillGrad. This ratio provides a descriptive measure of how effectively interaction experience is converted into validation improvement during skill evolution.

\subsection{Localized Replay Workload}
\label{app:localized_replay_workload}

Table~\ref{tab:localized_replay_workload} compares trigger-range replay with full-trajectory replay. \method{} reconstructs the replay state by copying the source prefix and executes only the target trigger range instead of replaying the complete trajectory. The audit covers all executed ranges from the 18 complete formal runs used in the mechanism analysis.

For range $r$, let $F_r$ denote the number of steps in the complete recorded source trajectory and $C_r$ denote the number of steps executed by \method{} during localized replay. We define the replay workload reduction as
\begin{equation}
  R_{\mathrm{step}}
  =1-\frac{\sum_r C_r}{\sum_r F_r}.
  \label{eq:localized_replay_reduction}
\end{equation}

\begin{table*}[t]
  \caption{Replay workload comparison between trigger-range replay and full-trajectory replay.}
  \label{tab:localized_replay_workload}
  \centering
  \small
  \setlength{\tabcolsep}{7.0pt}
  \renewcommand{\arraystretch}{1.08}
   \begin{tabular*}{\textwidth}{@{\extracolsep{\fill}}lrrrrrr@{}}
    \toprule
    Replay stage & Ranges &
    \shortstack{Full-trajectory\\replay steps} &
    \shortstack{Localized\\replay steps} &
    \shortstack{Copied\\prefix steps} &
    \shortstack{Reduction} \\
    \midrule
    Evidence-window replay & 2,505 & 53,242 & 18,007 & 20,714 & 66.18\% \\
    Post-rejection replay  &   207 &  3,863 &  1,205 &  1,629 & 68.81\% \\
    \bottomrule
  \end{tabular*}
\end{table*}

Trigger-range replay reduces the executed replay steps from 53,242 to 18,007 for evidence-window replay and from 3,863 to 1,205 for post-rejection replay, corresponding to reductions of 66.18\% and 68.81\%, respectively. These results show that \method{} avoids replaying trajectory regions beyond the evidence-supported ranges.

\section{Case Studies}
\label{app:case_studies}
\subsection{Replay-Guided Behavioral Verification of Edits}
\label{app:trace_cases}

\begin{figure*}[h]
    \centering
    \includegraphics[width=0.96\textwidth]{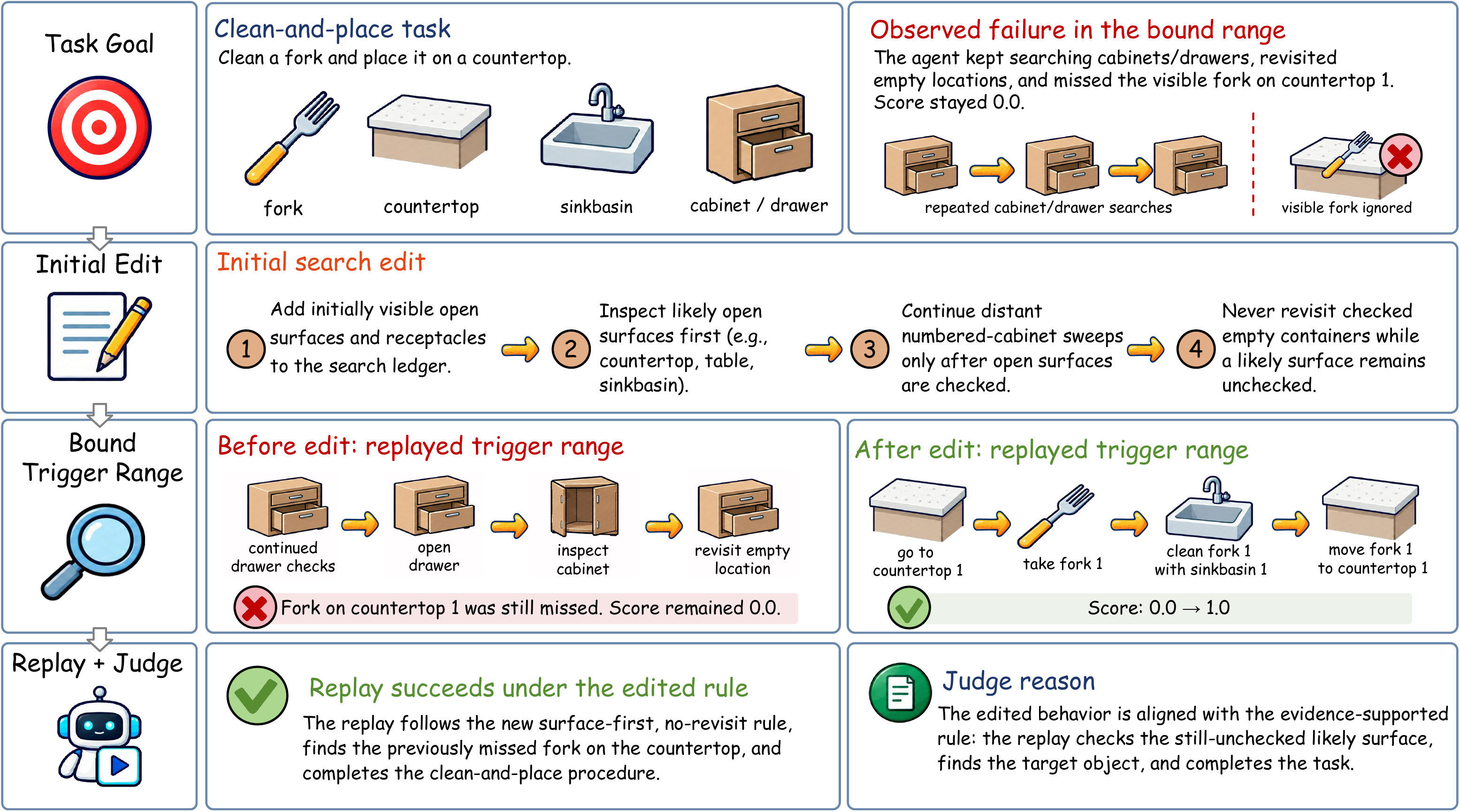}
    \caption{
    \textbf{Acceptance.}
    The Evidence-supported search edit changes behavior on the bound replay
    range, discovers the previously missed target, and completes the task,
    leading to \texttt{accept} decision.
    }
    \label{fig:case_accept}
\vspace{-5pt}
\end{figure*}

Following Section~\ref{sec:edit_verification}, each candidate edit $u$ is verified using its supporting cards $\mathcal{C}(u)$ and trigger ranges $\mathcal{G}(u)$. Replay reconstructs the source execution context and runs the Action Agent under $S_e^W\oplus u$. The evaluator compares the source and replayed behaviors to determine whether the intended correction is achieved without introducing new failures. We present four representative outcomes: \texttt{accept}, \texttt{reject}, \texttt{reflect}$\rightarrow$\texttt{accept}, and \texttt{reflect}$\rightarrow$\texttt{reject}, illustrating direct verification and feedback-guided refinement.

\paragraph{Case 1: Acceptance.}
Figure~\ref{fig:case_accept} illustrates an ALFWorld task requiring the agent to clean a fork and place it on a countertop. In the source trigger range, the agent repeatedly searches cabinets and drawers, revisits empty locations, and overlooks the fork on countertop 1. The supporting evidence therefore identifies a search-order problem: a likely open surface remains unchecked while previously inspected containers are revisited.
The candidate edit translates this observation into reusable guidance: record initially visible surfaces and receptacles, inspect likely open surfaces before distant cabinets, and avoid revisiting checked empty containers while plausible locations remain unexplored. Replaying the same source context under the edited skill produces the intended behavioral change. The agent visits countertop 1, retrieves the fork, cleans it using sinkbasin 1, and returns it to the countertop, increasing the score from 0.0 to 1.0.
The evaluator assigns \texttt{accept} without reflection because replay demonstrates both compliance with the proposed search rule and successful task completion. This case illustrates how evidence-linked replay tests whether an edit resolves the behavior that motivated it. Acceptance admits the edit to $\mathcal{U}_e^{+}$ for consolidation, while incorporation into the Validated Skill remains subject to global validation.

\begin{figure*}[h]
    \centering
    \includegraphics[width=0.96\textwidth]{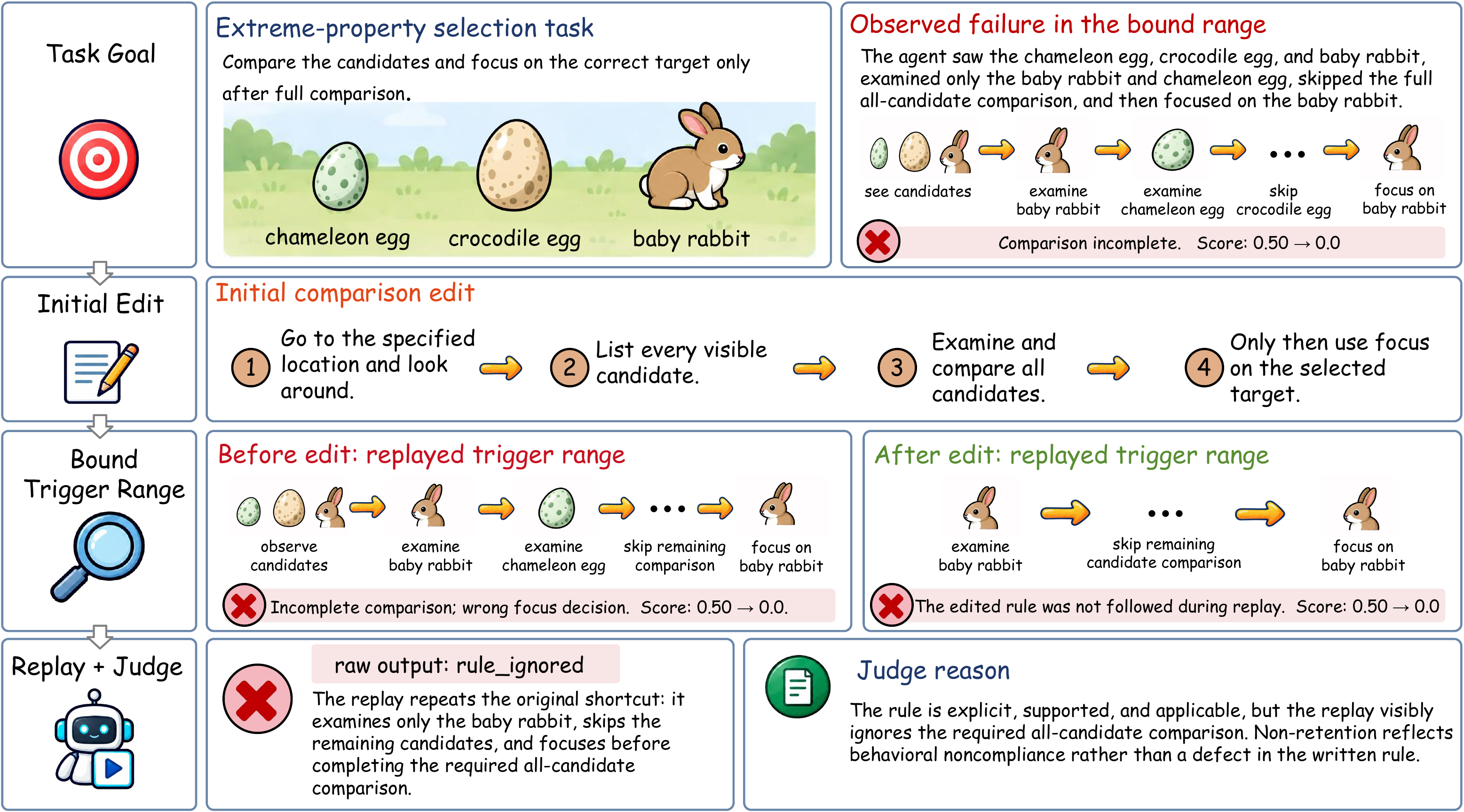}
  \caption{\textbf{Rejection due to behavioral noncompliance.} Replay skips the comparison required by the edit, resulting in \texttt{reject}.}
  \vspace{-15pt}
    \label{fig:case_reject}
\end{figure*}

\paragraph{Case 2: Rejection due to behavioral noncompliance.}
Figure~\ref{fig:case_reject} presents an extreme-property selection task requiring the agent to compare all visible candidates before focusing on the correct target. The source trajectory identifies a chameleon egg, a crocodile egg, and a baby rabbit, but examines only the baby rabbit and chameleon egg before focusing on the baby rabbit. The supporting evidence thus exposes an incomplete comparison that leaves one candidate unexamined.
The candidate edit specifies a complete procedure: reach the designated location, list every visible candidate, examine and compare all candidates, and only then focus on the selected target. However, replay from the source context under $S_e^W\oplus u$ shows that the agent examines only the baby rabbit, skips the remaining comparisons, and again focuses on the baby rabbit. The score remains 0.0, and the intended procedural correction is not realized.
The evaluator assigns \texttt{reject} because the replay ignores an explicit, applicable rule and provides no localized defect in the written edit to guide reflection. This case illustrates why evidence grounding alone does not establish behavioral effectiveness: an edit can address the observed failure in text while leaving execution uncorrected. The edit is excluded from $\mathcal{U}_e^{+}$, while its supporting cards remain available for subsequent proposals according to their lifecycle states.

\begin{figure*}[h]
    \centering
    \includegraphics[width=0.96\textwidth]{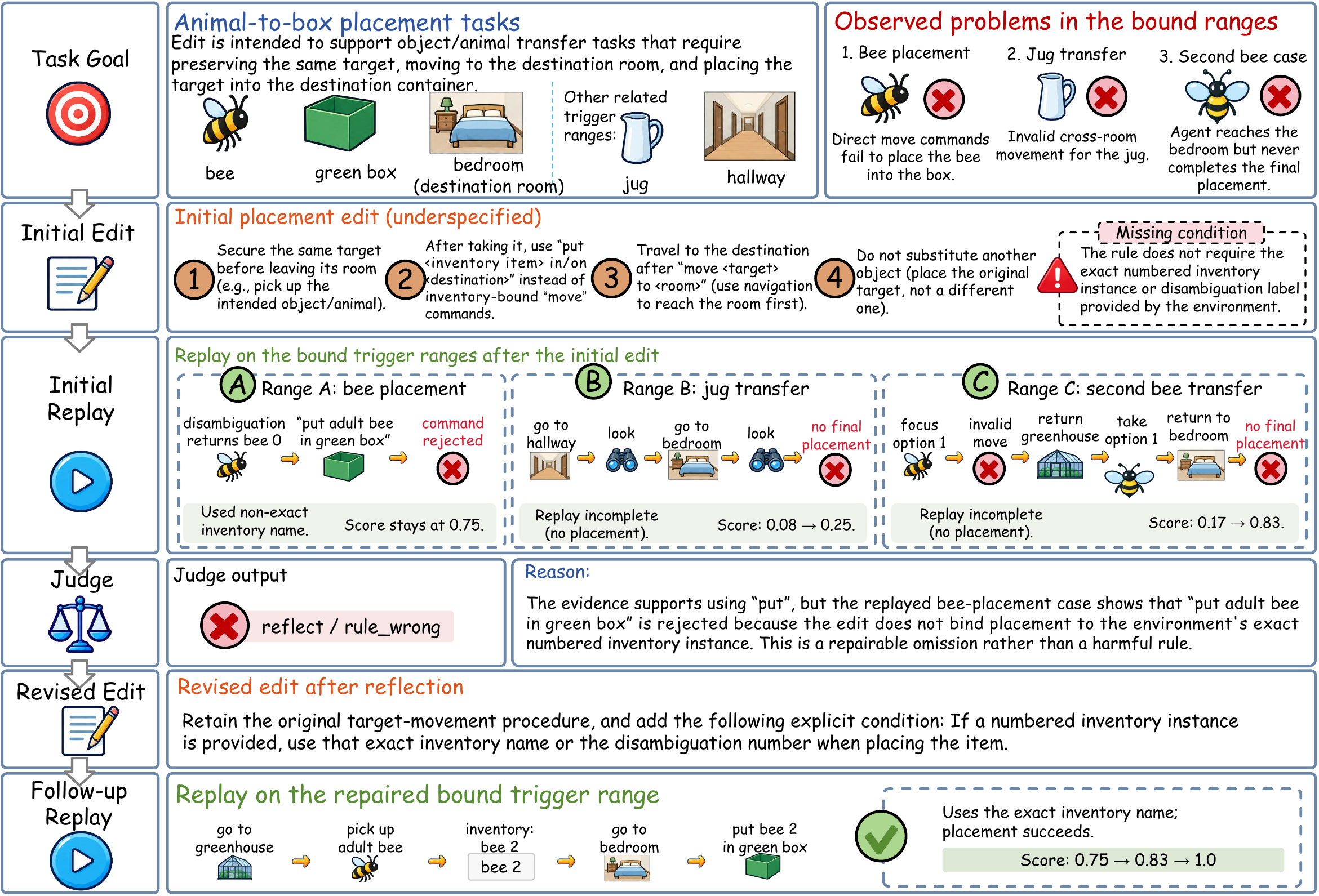}
    \caption{
    \textbf{Reflection followed by acceptance.}
    Initial replay exposes a missing exact-instance binding requirement in an
    otherwise useful placement rule. Reflection repairs this omission, and the
    revised edit subsequently completes the placement successfully, resulting
    in an \texttt{accept} decision.
    }
    \label{fig:case_reflect_accept}
\end{figure*}

\paragraph{Case 3: Reflection followed by acceptance.}
Figure~\ref{fig:case_reflect_accept} illustrates how replay feedback guides a localized revision. The supporting cards describe transfer failures involving bees and a jug, including invalid movement commands and reaching the destination without completing placement. The initial edit requires preserving the original target, navigating to the destination room, and using a placement command to put the held object into the destination container.
The initial replay exposes a missing constraint in this procedure. In the bee-placement range, the agent issues ``put adult bee in green box,'' which is rejected because the command does not use the exact numbered inventory instance. The score remains at 0.75. The other two ranges also end without placement, despite partial progress. The evaluator therefore assigns \texttt{reflect}: the evidence supports the placement procedure, but the edit leaves object identification underspecified.
Using this feedback, the editor produces $u'$ while retaining $\mathcal{C}(u)$ and the original transfer procedure. The revision explicitly requires the exact numbered inventory name when placing an item, rather than a generic name or a disambiguation number. In the follow-up replay, the agent checks its inventory, identifies ``bee 2,'' and issues ``put bee 2 in green box.'' Placement succeeds and the score reaches 1.0, yielding \texttt{accept}. This case demonstrates how replay identifies an actionable omission, directs a targeted repair, and verifies the revised edit before admitting it to $\mathcal{U}_e^{+}$.

\begin{figure*}[h]
    \centering
    \includegraphics[width=0.96\textwidth]{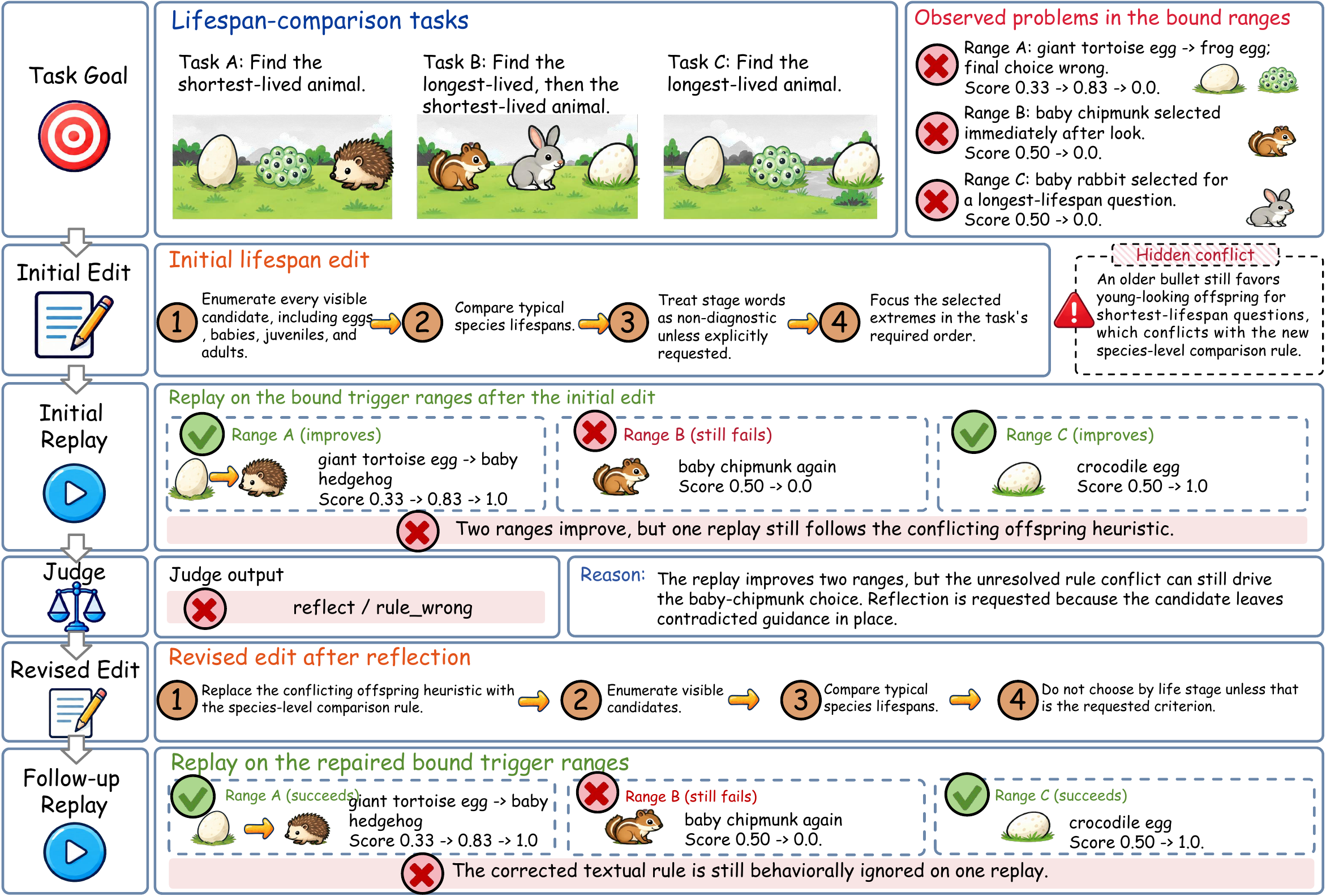}
    \caption{
    \textbf{Reflection followed by rejection.}
    Reflection removes the conflicting life-stage heuristic and repairs the
    written lifespan rule, but the corrected instruction remains ineffective on
    one replay range. The revised edit is therefore not retained.
    }
    \label{fig:case_reflect_reject}
\end{figure*}

\paragraph{Case 4: Reflection followed by rejection.}
Figure~\ref{fig:case_reflect_reject} presents lifespan-comparison tasks in which the agent must select the shortest- or longest-lived animal. The supporting evidence reveals incorrect choices involving eggs and young animals. The initial edit instructs the agent to enumerate all visible candidates, compare typical species lifespans, and disregard life-stage terms unless explicitly requested. However, an existing heuristic still favors young-looking offspring for shortest-lifespan questions, leaving conflicting guidance in the skill.
Initial replay improves ranges A and C: the agent correctly selects the baby hedgehog and crocodile egg, respectively, and both scores reach 1.0. Range B still fails because the agent again selects the baby chipmunk, reducing the score from 0.50 to 0.0. The evaluator assigns \texttt{reflect} because the unresolved conflict provides a specific target for revision. The editor replaces the offspring heuristic with the species-level comparison rule and explicitly prohibits selection by life stage unless the task requests it.
Follow-up replay under $S_e^W\oplus u'$ preserves the successes in ranges A and C, but range B repeats the same incorrect choice. The evaluator therefore assigns \texttt{reject}, and the revised edit is excluded from $\mathcal{U}_e^{+}$. This case illustrates why reflection must be followed by renewed behavioral verification: removing a textual conflict does not ensure that execution follows the corrected rule, and improvements in some supporting contexts do not resolve a persistent failure in another.

\subsection{Cross-Task Support for a Shared Evidence Card}
\label{app:cross_task_evidence}

A representative ALFWorld case illustrates how a single Evidence Card links replayable trigger ranges from multiple tasks that exhibit the same execution requirement. As summarized in Table~\ref{tab:cross_task_trace}, the Card is supported by four successful trajectories involving different objects, transformations, and destination receptacles.

\begin{table}[!htbp]
  \caption{Four task-specific trigger ranges linked to a shared Evidence Card.}
  \label{tab:cross_task_trace}
  \centering
  \scriptsize
  \setlength{\tabcolsep}{3.5pt}
  \renewcommand{\arraystretch}{1.08}
  \begin{tabularx}{\columnwidth}{@{}l>{\raggedright\arraybackslash}X@{}}
    \toprule
    Task & Linked trigger range (intermediate navigation omitted) \\
    \midrule
    \texttt{Task1} & take potato 1; heat it with microwave 1; move potato 1 to diningtable 1. \\
    \texttt{Task2} & take bowl 2; cool it with fridge 1; move bowl 2 to cabinet 1. \\
    \texttt{Task3} & take bread 2; heat it with microwave 1; move bread 2 to diningtable 1. \\
    \texttt{Task4} & take fork 1; clean it with sinkbasin 1; move fork 1 to countertop 1. \\
    \midrule
    Shared rule & Maintain the exact acquired instance through transformation and final placement; do not substitute another visible instance. \\
    \bottomrule
  \end{tabularx}
\end{table}

Although the four tasks differ in their manipulated objects, transformations, and target receptacles, their linked trigger ranges support the same correction: the agent must preserve the identity of the acquired object instance throughout execution. The shared Evidence Card consolidates this cross-task support into reusable behavioral guidance instead of representing each trajectory as an independent task-specific rule.

\subsection{Cross-Epoch Correction of a Tracked Edit}
\label{sec:edit_correction}

\begin{figure*}[h]
    \centering
    \includegraphics[width=\textwidth]{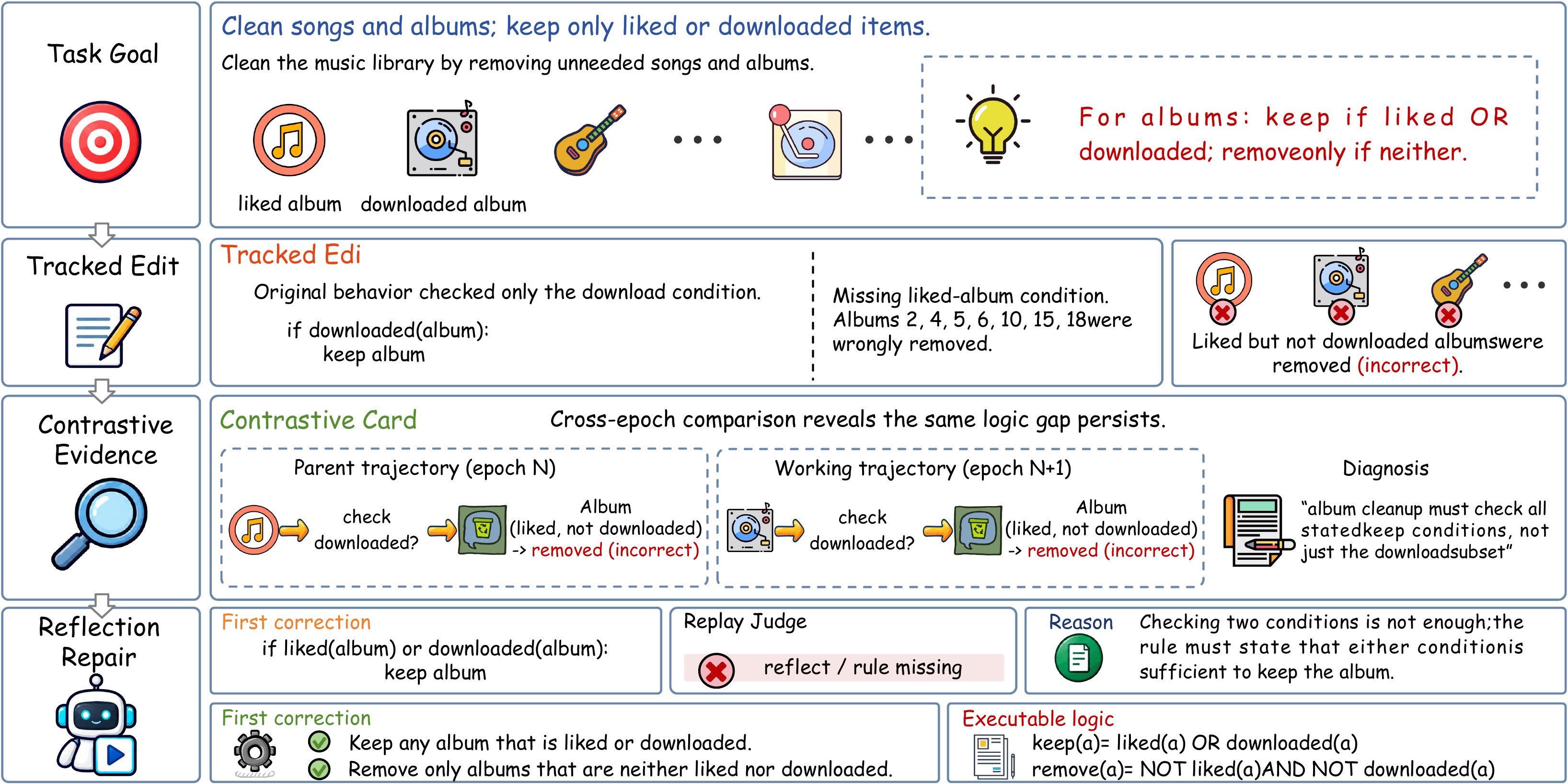}
    \caption{Cross-epoch evidence reveals a persistent edit deficiency, guiding correction through reflection and replay.}
    \label{fig:spotify_case}
\end{figure*}

Figure~\ref{fig:spotify_case} illustrates how \method{} revisits an existing edit when its deficiency persists across epochs. In this AppWorld task, the agent must clean a music library while preserving albums that are liked or downloaded. The tracked edit checks only download status, causing liked but non-downloaded albums to be removed. Its continued use therefore preserves an incomplete rule that requires further correction. The evidence links make this persistent defect traceable. As described in Section~\ref{sec:edit_construction}, the edit belongs to $\mathcal{Q}_e$, and its supporting cards $\mathcal{C}(u)$ identify the task whose adjacent-epoch trajectories enter $\mathcal{R}_e^{\mathrm{cmp}}$. Both trajectories exhibit the same erroneous removal pattern. The resulting Contrastive Evidence Card associates this recurring failure with the tracked edit and proposes checking all stated preservation conditions. Cross-epoch propagation thus supplies evidence for reassessing guidance already used by the Working Skill.

The first correction requires checking both liked and downloaded status, but replay feedback identifies a remaining ambiguity: either condition alone must be sufficient for preservation. Reflection makes this requirement explicit: keep an album if it is liked or downloaded, and remove it only if it is neither. The revised edit passes replay verification and is consolidated into a candidate revision that subsequently passes global validation. This case connects the three Method phases: retained evidence enables targeted cross-epoch diagnosis, synthesis revises the tracked edit, and replay verifies the correction before global validation incorporates it into the Validated Skill.
    
\section{Core Method Prompts}
\label{app:prompts}
\label{app:prompts}

This appendix presents five prompt templates supporting the three phases of \method{} in Section~\ref{sec:method}: Replayable Evidence Card extraction, card grounding verification, evidence-grounded edit synthesis, replay-guided edit evaluation, and contrastive evidence extraction. These templates specify how execution observations are recorded as evidence, translated into localized edits, and used to assess or revisit the resulting guidance.

The templates retain their original Markdown formatting. Run-specific inputs, including skills, trajectories, cards, candidate edits, and evaluator feedback, are supplied separately as user messages. The descriptions below relate each template to the notation and operations used in the Method section.

\subsection{Replayable Evidence Card Extraction}

The LLM evidence extractor $f^{\mathrm{ext}}$ receives the Working Skill $S_e^W$ and training trajectories $\mathcal{R}_e$. It examines successful and failed interactions to identify reusable corrections or improvements to procedures, preconditions, and recovery guidance. Each proposal is recorded as a Replayable Evidence Card $C_i=\langle i,\mathcal{G}_i,d_i\rangle$, linking the proposed correction $d_i$ to supporting trigger ranges $\mathcal{G}_i$. The resulting cards form $\mathcal{C}_e^{\mathrm{traj}}$ and provide the execution references needed for subsequent edit synthesis and replay.

\begin{mdframed}[
    backgroundcolor=gray!5,
    linecolor=gray!35,
    linewidth=0.8pt,
    roundcorner=3pt,
    skipabove=8pt,
    skipbelow=10pt
]
\begin{lstlisting}[
    style=promptstyle,
    backgroundcolor=\color{gray!5}
]
You analyze a minibatch of train trajectories and propose concrete edits to the current skill.md.

Input contains the full skill, source_type, max_proposed_edits, and compact trajectories. Each trajectory shows its task, outcome, initial observation, every action, and the environment feedback after that action. It may also contain per-step score transitions. Read every trajectory before proposing edits.

Prefer reusable mechanisms that recur in the minibatch. A single failed trajectory may justify a narrow edit only when visible feedback directly contradicts the current skill or exposes a concrete missing executable procedure. Never memorize an answer, product, location, object placement, or other incidental case detail.

Failure types:
- rule_missing: skill.md lacks a needed transferable decision rule.
- rule_wrong: skill.md gives misleading, too-specific, stale, or contradicted guidance.
- rule_ignored: skill.md has a useful rule but the trajectory violates it or needs stronger action-level wording.
- invalid_action: repeated invalid or non-executable actions.
- exploration_gap: missed relevant states, locations, objects, tools, options, readings, or disambiguation.
- experiment_error: poor test or interaction sequence, missing observation, or missing verification.
- premature_finish: stopping or submitting before the goal is verified.
- loop: repeated actions or observations without progress.
- other: none of the above.

Return only valid JSON:
{
  "evidence": [
    {
      "experience_type": "failure_reflection | success_experience",
      "failure_type": "rule_missing | rule_wrong | rule_ignored | invalid_action | exploration_gap | experiment_error | premature_finish | loop | other | null",
      "pattern": "short reusable problem or success mechanism",
      "proposed_edit": {
        "op": "append | insert_after | insert_after_section | replace | delete",
        "target": "exact target text or heading; omit or null for append",
        "content": "markdown content; omit or empty for delete"
      },
      "trigger_ranges": [
        {"task_id": "task id from the input", "start": 0, "end": 0}
      ]
    }
  ]
}

Rules:
- Return at most max_proposed_edits entries; an empty evidence list is valid.
- trigger_ranges must be non-empty and use visible task IDs and step numbers.
- Each proposed_edit must be directly supported by its trigger ranges.
- One entry may cite multiple tasks only when they support the same edit.
- pattern must be short and mechanism-level; proposed_edit must be concise and actionable.
- Judge steps from action, post-action feedback, optional score change, done status, and final outcome together. Positive score change supports progress and negative change supports regression, but zero or unavailable score alone does not make a setup, navigation, observation, or verification step bad.
- Repeated zero-progress steps support a loop only when actions or feedback also repeat without useful state change.
- rule_wrong should replace, delete, soften, or generalize contradicted guidance rather than append a competing rule.
- Do not output supporting_tasks, before_segments, trigger_progress, source_type, support_count, merge_level, evidence IDs, or commentary outside JSON. The system materializes those fields from real train trajectories.

===== FAILURE-TRAJECTORY ADDENDUM =====
Analyze failed trajectories only. Set experience_type to failure_reflection and include a non-null failure_type. Prefer systematic failures, but retain a narrow directly observed skill contradiction. Use visible outcome and post-action feedback as authoritative evidence.

===== SUCCESS-TRAJECTORY ADDENDUM =====
Analyze successful trajectories only. Set experience_type to success_experience and omit failure_type. Extract reusable successful procedures that add operational detail beyond skill.md. Preserve necessary setup, navigation, observation, and verification steps. Do not discard a stable task-family procedure merely because unrelated trajectories do not share it.
\end{lstlisting}
\end{mdframed}

\subsection{Evidence Card Grounding Verification}

The grounding verifier checks a proposed card against its cited trajectory spans. Its purpose is to establish whether the recorded observations support the proposed correction and whether the evidence references locate the relevant behavior. This check supports the grounding step in Evidence-Grounded Edit Synthesis. It does not establish that an edit derived from the card will change execution as intended, which is assessed separately through replay.

\begin{mdframed}[
    backgroundcolor=gray!5,
    linecolor=gray!35,
    linewidth=0.8pt,
    roundcorner=3pt,
    skipabove=8pt,
    skipbelow=10pt
]
\begin{lstlisting}[
    style=promptstyle,
    backgroundcolor=\color{gray!5}
]
You verify whether one system-materialized EvidenceCard is factually supported by its real train trajectory ranges.

The card contains source_type, experience_type, optional failure_type, pattern, proposed_edit, trigger_ranges, supporting_tasks, and before_segments generated by the system from train rollout data.

Return only valid JSON:
{
  "decision": "keep | drop",
  "reason": "brief reason grounded in before_segments, task outcome, and current skill.md"
}

Keep when the visible initial state, actions, post-action feedback, optional progress signals, and task outcome support the pattern and proposed edit. For failure evidence, failure_type must match the visible mechanism. Drop invalid ranges, malformed edits, unsupported claims, wrong labels, case memorization, or unjustified broad defaults.

Do not drop evidence merely because skill.md already covers it, another card is redundant, its priority is low, or its support is narrow. Coverage, deduplication, and scope handling happen later. Keep the reason concise.
\end{lstlisting}
\end{mdframed}

\subsection{Evidence-Grounded Edit Synthesis}

The LLM editor $f^{\mathrm{edit}}$ receives the Working Skill $S_e^W$ and an Evidence Window $\mathcal{B}_k$ containing active cards with compatible contexts and correction intents. It refines or consolidates their proposals into localized candidate edits, preserving the persistent identifiers of the cards supporting each edit. These associations define $\mathcal{C}(u)$ and determine the trigger ranges $\mathcal{G}(u)$ used for verification. When contrastive evidence targets an existing edit, it guides a correction to that guidance.

\begin{mdframed}[
    backgroundcolor=gray!5,
    linecolor=gray!35,
    linewidth=0.8pt,
    roundcorner=3pt,
    skipabove=8pt,
    skipbelow=10pt
]
\begin{lstlisting}[
    style=promptstyle,
    backgroundcolor=\color{gray!5}
]
You are a Skill Edit Coverage Merger for a reusable markdown skill document.

Input contains current skill.md, one source_type, and multiple lightweight verified EvidenceCards. Each card contains evidence_id, source_type, optional failure_type, pattern, proposed_edit, support_count, merge_level, source_minibatch_id, optional task_families, and optional trigger_progress summaries, but no raw trajectory.

Task: merge, deduplicate, and repair conflicts among the concrete proposed_edit values. Cover every useful EvidenceCard with an edit, or explicitly omit it with a reason. Do not invent edits unrelated to cited proposed_edit values.

Guidelines:
- This is coverage merging, not epoch ranking or final budgeting. Never silently drop cards because there are many.
- All cards belong to one semantic window; compare their edit target and content before writing.
- Write reusable decision procedures, not answer, product, object, or location memorization.
- Cite only EvidenceCards that directly support an edit.
- Keep local or single-case support narrow and conditional.
- source_type=success is reusable positive procedure evidence, not low-priority filler.
- source_type=failure supports corrections, guardrails, or recovery procedures.
- rule_wrong must replace, delete, soften, or generalize misleading guidance. Do not leave the contradicted rule in force beside a new competing rule.
- Place edits in an exact matching section when possible. If none fits, add a concise mechanism-level subsection without rewriting unrelated content.
- Return empty edits only when every card is explicitly omitted as already covered, redundant, conflicting, unsafe, unsupported, or too local.

Return only valid JSON:
{
  "reasoning": "brief explanation of consolidation decisions",
  "omitted_evidence": [
    {"evidence_id": "E000009", "reason": "already_covered | redundant | conflicting | unsafe | unsupported | too_local"}
  ],
  "edits": [
    {
      "op": "append | insert_after | insert_after_section | replace | delete",
      "target": "exact markdown heading or target text; omit or null for append",
      "content": "markdown content to add or replace; omit or empty for delete",
      "evidence_ids": ["E000001", "E000008"]
    }
  ]
}

For insert_after_section, insert_after, replace, and delete, target must exactly match current skill.md. content must be markdown. Evidence IDs are metadata and must not appear in markdown content. Do not output text outside the JSON object.
\end{lstlisting}
\end{mdframed}

\subsection{Replay-Guided Edit Verification}

The LLM evaluator $f^{\mathrm{eval}}$ assesses a candidate edit $u$ using its supporting cards $\mathcal{C}(u)$ and source--replay segment pairs $\mathcal{T}_u$. It examines whether execution under $S_e^W\oplus u$ achieves the intended correction without introducing new failures. The output comprises a decision $r_u$ and feedback $h_u$: \texttt{accept} indicates replay support, \texttt{reflect} identifies a localized deficiency that can guide revision, and \texttt{reject} excludes an unsupported edit. For reflection, the feedback guides $f^{\mathrm{edit}}$ in producing a revised edit that undergoes another replay and evaluation.

\begin{mdframed}[
    backgroundcolor=gray!5,
    linecolor=gray!35,
    linewidth=0.8pt,
    roundcorner=3pt,
    skipabove=8pt,
    skipbelow=10pt
]
\begin{lstlisting}[
    style=promptstyle,
    backgroundcolor=\color{gray!5}
]
You are a Skill Edit Judge for reusable markdown skill edits.

Input contains current skill.md, one candidate edit, cited EvidenceCards with source_type, experience_type, optional failure_type, pattern, supporting_tasks, concrete proposed edits, and real before_segments, plus optional replay_evidence showing after_segments. Decide whether to accept the edit.

Return only valid JSON:
{
  "decision": "accept | reflect | reject",
  "replay_failure_type": "rule_missing | rule_wrong | rule_ignored | null",
  "reason": "brief reason grounded in linked EvidenceCards, replay evidence, and current skill.md"
}

`replay_failure_type` describes the candidate skill after applying the edit, not the original EvidenceCard:
- rule_ignored: the candidate contains an explicit, correct, executable rule; replay reaches a state where it applies; and the replay action visibly violates it. Generic advice such as "be systematic" or "verify carefully" is insufficient.
- rule_missing: the candidate omits or underspecifies a condition, step, procedure, stop condition, exact action form, or verification required by linked evidence.
- rule_wrong: the candidate itself contains contradicted, conflicting, or harmful guidance, or replay follows that guidance and fails because of it.
- null: replay supports the candidate, replay is inconclusive for reasons unrelated to the edit, or the candidate should be rejected for support, safety, or reusability rather than one of the three skill failures.

Decision mapping is mandatory: rule_ignored -> reject; rule_missing or rule_wrong -> reflect; null -> accept or reject based on evidence. Candidate-level attribution priority is rule_wrong, then rule_missing, then rule_ignored, then null. Use rule_ignored only when no replay reveals a candidate rule that is wrong or missing. A prefix mistake, unrelated invalid action, stochastic noncompliance, a short replay that does not finish, or final failure alone does not invalidate an otherwise supported candidate. Input may set `reflection_allowed=false`; in that case do not return reflect and reject a rule_missing/rule_wrong candidate.

Accept only when the candidate is supported by cited proposed_edit values, fills or repairs a real skill need, and states a reusable executable procedure. New headings must organize a well-supported mechanism. Reject weak, redundant, verbose, conflicting, overgeneralized, case-memorizing, or unsupported edits. Reject broad defaults from local evidence and rule_wrong edits that merely append a new rule while leaving contradicted text unresolved.

Replay rule: success_experience should preserve useful behavior; failure_reflection should improve or avoid the mistake. Mixed edits must preserve successes and improve failures. Replay need not finish the task inside a short range. Reject regression only when the candidate rule caused it; when a correct explicit rule was applicable and the replay action violated it, use rule_ignored instead. Judge markdown content; evidence IDs are metadata.
\end{lstlisting}
\end{mdframed}

\subsection{Contrastive Evidence Card Extraction}

The LLM evidence extractor $f^{\mathrm{ext}}$ receives the Working Skill $S_e^W$, the Tracked Edits $\mathcal{Q}_e$ with their supporting cards, and the adjacent-epoch trajectory pairs in $\mathcal{R}_e^{\mathrm{cmp}}$. The tracked edits include live provisional edits and edits newly incorporated into the Validated Skill during the preceding epoch transition. Their evidence links identify the training tasks to compare. The extractor identifies persistent deficiencies or newly exposed problems and records proposed corrections as $\mathcal{C}_e^{\mathrm{con}}$. These cards use the same representation as trajectory-derived cards, while retaining any association with an existing edit for targeted revision.

\begin{mdframed}[
    backgroundcolor=gray!5,
    linecolor=gray!35,
    linewidth=0.8pt,
    roundcorner=3pt,
    skipabove=8pt,
    skipbelow=10pt
]
\begin{lstlisting}[
    style=promptstyle,
    backgroundcolor=\color{gray!5}
]
You compare two saved train trajectories for the same task before and after a provisional skill edit.

Input contains the dataset, comparison_type, the relevant provisional edits, the parent and working skill revisions, the current working skill, and compact parent/working train trajectories. It never contains validation or test data.

Return only valid JSON:
{
  "decision": "support | evidence | no_evidence",
  "failure_type": "rule_missing | rule_wrong | rule_ignored | null",
  "reason": "brief causal comparison grounded in the rule, applicable state, and action difference",
  "evidence": {
    "pattern": "short reusable mechanism",
    "proposed_edit": {
      "op": "append | insert_after | insert_after_section | replace | delete",
      "target": "exact target text or heading; omit or null for append",
      "content": "markdown content; omit or empty for delete"
    },
    "trigger_range": {"task_id": "the compared task id", "start": 0, "end": 0}
  }
}

Rules:
- A hard-success change alone does not establish causality. Compare the relevant rule, whether its precondition was reached, and the actual actions and feedback.
- For failure -> success, return support only when the origin rule applied and the action difference shows that following it caused or materially enabled success. Otherwise return no_evidence. Never generate a duplicate edit.
- For success -> failure, return correction evidence only when an origin edit caused or exposed a concrete wrong or missing procedure.
- For failure -> failure, return evidence only for an explicit missing or wrong procedure. A merely ignored correct rule is no_evidence with failure_type=rule_ignored.
- For success -> success, return no_evidence.
- rule_missing means the working skill lacks a necessary executable condition or step. rule_wrong means an origin edit contains incorrect or conflicting guidance. rule_ignored means a correct explicit rule applied but the agent violated it.
- Any evidence must be a narrow correction of the listed origin edits. Do not add unsupported APIs, parameters, answers, product IDs, object locations, entity IDs, credentials, or private values.
- trigger_range must reference visible steps from the working trajectory. Do not mention validation, test data, optimization, or EvidenceCards in markdown content.
\end{lstlisting}
\end{mdframed}

\end{document}